\documentclass[accepted]{uai2022} 

\usepackage[british]{babel}

\usepackage{natbib} 
\usepackage{mathtools} 
\usepackage{booktabs} 
\usepackage{tikz} 

\usepackage{graphicx}
\usepackage{algorithm}
\usepackage{algorithmic}
\usepackage{amssymb}

\newcommand\argmax{\mathop{\rm arg\,max}}

\newcommand {\defn} {\triangleq}
\newcommand \Reals {\ensuremath{\mathbb{R}}}

\renewcommand \Pr {\mathop{\mbox{\ensuremath{\mathbb{P}}}}\nolimits}
\newcommand \pol {\pi}

\newcommand \bel {\beta}
\newcommand \mdp {\mu}

\newcommand \dd {\, \mathrm{d}}
\let\Pr\relax
\newcommand \Pr {\mathbb{P}}
\newcommand \UE {U_{\alpha_2}^E}
\newcommand \UA {U_{\alpha_1}^A}

\DeclareMathAlphabet{\mathpzc}{OT1}{pzc}{m}{it}

\newcommand \Beta {{\mathpzc{Beta}}}

\if 1

\else

\fi

\usepackage{caption,subcaption}
\usepackage{amsmath,array, amssymb, amsthm}
\usepackage{tikz}

\newif\ifsinglecol
\singlecolfalse

\newtheorem{example}{Example} 
\newtheorem{theorem}{Theorem}

\newtheorem{remark}[theorem]{Remark}

\newtheorem{definition}[theorem]{Definition}

\newtheorem{axiom}[theorem]{Axiom}

\title{SENTINEL: Taming Uncertainty with Ensemble based Distributional Reinforcement Learning}

\author[1, 2]{\href{mailto:<hannese@chalmers.se>?Subject=SENTINEL: Taming Uncertainty with Ensemble based Distributional Reinforcement Learning}{Hannes Eriksson}{}}
\author[3, 4]{Debabrota Basu}
\author[1]{Mina Alibeigi}
\author[2,5]{Christos Dimitrakakis}
\affil[1]{%
    Zenseact AB\\
    Gothenburg, Sweden
}
\affil[2]{%
    Chalmers University of Technology\\
    Gothenburg, Sweden
}
\affil[3]{%
    Scool\\
    INRIA Lille-Nord Europe\\
    Lille, France
  }
  
\affil[4]{%
    CRIStAL,
    CNRS,
    Lille, France
  }
  
\affil[5]{%
    University of Neuchatel, Switzerland
    and
    University of Oslo, Norway
  }
  
\begin{document}
\maketitle

\begin{abstract}
    In this paper, we consider risk-sensitive sequential decision-making in Reinforcement Learning (RL). 
    Our contributions are two-fold. First, we introduce a novel and coherent quantification of risk, namely composite risk, which quantifies the joint effect of aleatory and epistemic risk during the learning process.
    Existing works considered either aleatory or epistemic risk individually, or as an additive combination.
    We prove that the additive formulation is a particular case of the composite risk when the epistemic risk measure is replaced with expectation.
    Thus, the composite risk is more sensitive to both aleatory and epistemic uncertainty than the individual and additive formulations.
    We also propose an algorithm, SENTINEL-K, based on ensemble bootstrapping and distributional RL for representing epistemic and aleatory uncertainty respectively. The ensemble of K learners uses Follow The Regularised Leader (FTRL) to aggregate the return distributions and obtain the composite risk.
    We experimentally verify that SENTINEL-K estimates the return distribution better, and while used with composite risk estimates, demonstrates higher risk-sensitive performance than state-of-the-art risk-sensitive and distributional RL algorithms.
\end{abstract}


\section{Introduction}\label{sec:intro}
Reinforcement Learning (RL) algorithms, with their recent success in games and simulated environments~\citep{mnih2015human}, have drawn interest for real-world and industrial applications~\citep{pan2017virtual,mahmood2018benchmarking}. 
In addition, since in RL the environment is by definition unknown to the agent, exploring it so as to improve performance and eventually obtain the optimal policy entails risks.
Although the risk is not an issue in simulation, it is important to consider risks when interacting in the real world~\citep{pinto17a,garcia2015comprehensive,DBLP:journals/corr/abs-1810-09126}.
In this paper, we employ a model-free approach that enables us both to efficient in terms of the amount of data needed, and to be flexible with respect to the risk metric the agent should consider when making decisions.
\begin{figure*}[t!]
	\centering
	\includegraphics[width=0.79\textwidth]{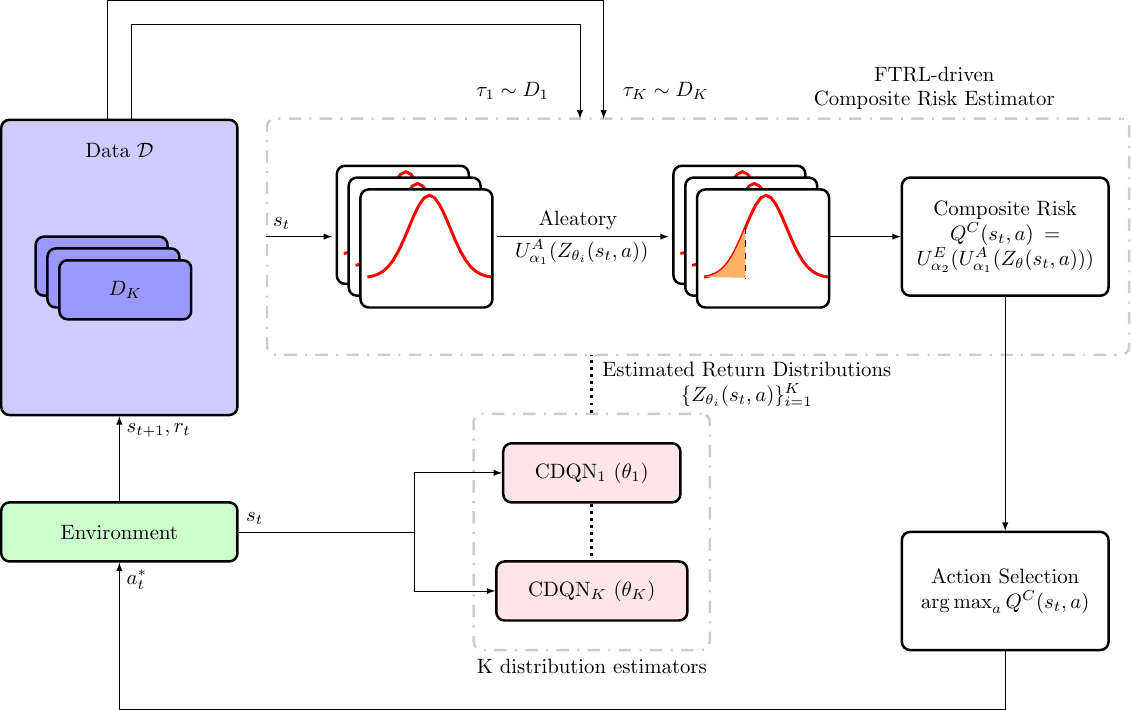}
	\caption{SENTINEL-K with FTRL-driven composite risk estimator and K CDQNs as return distribution estimators.}
	\label{fig:blockdiagram}
\end{figure*}

Risk sensitivity in reinforcement learning and Markov Decision Processes (MDPs) has sometimes been considered under a minimax formulation over plausible MDPs~\citep{satia:uncertain,HEGER1994105,tamar2014scaling}.
Alternative approaches include maximising a risk-sensitive statistic instead of the expected return~\citep{chow2014algorithms,tamar2015optimizing,clements2019estimating}.
In this paper, we focus on the second approach due to its flexibility.
Either approach requires estimating the uncertainty associated with the decision-making procedure.
This uncertainty includes both the inherent randomness in the model and the uncertainty due to imperfect information about the true model.
These two type of uncertainties are called \textit{aleatory} and \textit{epistemic} uncertainty respectively~\citep{der2009aleatory}.

In recent literature, researchers have either quantified epistemic and aleatory risks separately~\citep{mihatsch2002risk,eriksson2019epistemic} or considered an additive risk formulation where their weighted sum is minimised by an RL algorithm~\citep{clements2019estimating}. 

In this work, we propose a \textit{composite risk} formulation in order to accurately capture the combined effect of aleatory and epistemic uncertainty for decision-making in RL (Section~\ref{sec:risk}). Our composition of risks relies on \emph{coherent} risk measures, for which we show that their composition remains coherent. Our choice of focusing on coherent risk measures is also motivated by its extensive use and corresponding benefits in control theory~\cite{majumdar2017risk}, decision theory~\cite{pflug2016time}, and reinforcement learning theory~\cite[and references therein]{tamar2016sequential,ruszczynski2010risk}.


We incorporate composite risk measures within the Distributional RL (DRL) framework~\citep{bellemare2017distributional,tang2018exploration,rowland2019statistics}.
The DRL framework aims to model the distribution of returns of a policy for a given environment (Section~\ref{sec:drl}).
This highly expressive distributional representation allows us to both estimate appropriate risk measures and to incorporate them in final decision-making.
However, DRL approaches are typically limited to modelling aleatory uncertainty, with epistemic uncertainty due to partial information not being explicitly modelled in terms of the return distribution.
We us a bootstrapping~\citep{efron1985bootstrap} framework to represent epistemic uncertainty. Our framework, which we call SENTINEL-K, is illustrated in Figure~\ref{fig:blockdiagram}. At a high level, we use Categorical Deep Q Network (CDQN)~\citep{bellemare2017distributional} to model aleatory uncertainty and a bootstrapped ensemble for epistemic uncertainty. These can be used with any coherent measures and ensemble algorithm.

We discuss related work in Section~\ref{sec:related}. This is followed by some background on risk
measures, Markov decision processes, and DRL in
Section~\ref{sec:background}. SENTINEL-K is flexible enough to use any combination of
coherent risk measures for aleatory and epistemic risks, as we explain in Section~\ref{sec:risk}.  The algorithm is described in detail in Section~\ref{sec:algo}, with Section~\ref{sec:ensemble} and ~\ref{sec:ftrl} showing how the ensemble is created and its members weighted respectively.

 Section~\ref{sec:experiments} examines the performance of SENTINEL-K with a composite CVaR metric on a highway environment with $10$ cars. Our results show that our approach leads to fewer number of crashes than competing algorithms: Variational DQN (VDQN)~\citep{tang2018exploration}, CDQN~\citep{bellemare2017distributional}, total variance decomposition Uncertainty Aware-DQN (UA-DQN)~\citep{clements2019estimating}, as well as SENTINEL-K with additive CVaR estimate, which we used as an ablation test to showcase the importance of the using a coherent composite risk.
The supplementary material includes further experiments, showing that SENTINEL-K  features significantly improved estimates of return distributions, and shows that using FTRL for weighing the ensemble members measurably improves performance.

\section{Related Work}\label{sec:related}
For RL applications in the real world, such as for autonomous driving and robotics, \textit{risk-sensitive} RL approaches can avoid the negative consequences of excessive exploration that may lead to unsafe decisions in real-life.
This has initiated a spate of research efforts~\citep{howard1972risk,satia:uncertain, coraluppi1999risk, marcus1997risk, mihatsch2002risk,DBLP:journals/corr/abs-1810-09126} spanning five decades. But the majority of risk-sensitive RL papers~\citep{howard1972risk, coraluppi1999risk, marcus1997risk} focused on discrete state-space MDPs and either aleatory or epistemic risk.
We are interested in designing a general risk-sensitive framework applicable to any type of state space and risk.

Both \textit{aleatory} and \textit{epistemic} uncertainties are important for risk-sensitive RL.
The former expresses the \emph{randomness} inherent to the problem and the latter a \emph{lack of knowledge} about the problem. Aleatory risk-sensitivity in MDPs was first considered by~\citep{howard1972risk}, who introduced the idea of exponential utilities for the return.\footnote{Here, we use return to mean the total discounted reward} Epistemic uncertainty in MDPs was investigated by~\citep{satia:uncertain}, who provided game theoretic and Bayesian solution methods. Later works~\citep{coraluppi1999risk, marcus1997risk, mihatsch2002risk} extend risk-neutral methods to the risk-sensitive setting by using a non-linear utility~\citep{garcia2015comprehensive}. They consider aleatory risk-sensitive RL with exponential utility on the return \citep{mihatsch2002risk}.
Follow-up works~\citep{chow2014algorithms, chow2015riskconstrained} focus on scaling up these approaches. Other work on risk-sensitive RL focuses on CVaR~\citep{chow2014algorithms, tamar2015optimizing, chow2015risk}.
There have been recent works considering epistemic risk~\citep{eriksson2019epistemic}, wherein problem uncertainty is expressed in a Bayesian framework as a distribution over MDPs.
\cite{depeweg2018decomposition, clements2019estimating} intuitively incorporates both of these risks in decision making. 
\cite{depeweg2018decomposition} considers the risk in the per-step rewards obtained in a MDP while 
\cite{clements2019estimating} proposes to use the additive formulation of epistemic and aleatory risks. Both of them use variance, which is not a coherent measure~\citep{coherent_risk}. Unlike previous work, our methodology of composite risk also allows us to apply any pair of coherent risk measures\footnote{For example, CVaR, Wang risk measure~\citep{Wang2002ARM}, Standard Deviation (SD).} to aleatory and epistemic uncertainty.

We instead define a generalised composite risk measure that takes into account both epistemic and aleatory uncertainty, and their entangled effect. Coherence is important, as we show that for any two coherent risk measures the composite risk retains coherence. This gives a principled approach for combining different application-appropriate risk measures for epistemic and aleatory uncertainties.


To express aleatory uncertainty, we rely on a distributional RL method called CDQN, which incorporates highly expressive approximators to model continuous and multimodal return distributions. In addition, we leverage ensemble methods to express epistemic uncertainty. Ensemble methods have first been used in risk-neutral RL by for representing epistemic uncertainty in order to improve  exploration~\citep{dimitrakakis2006nearly,dimitrakakis2007ensembles}. This approach was later applied to MDPs by~\citet{osband2016deep}. On the other hand, \citet{wiering2008ensemble} used ensembles to combine policies instead. Ensembles have also been used to represent aleatory~\citep{fausser2015neural,  pacchiano2020optimism} uncertainty. Recently, \citep{depeweg2018decomposition,clements2019estimating} also use multiple Bayesian Neural Networks (BNNs) to estimate epistemic uncertainty. In the best of our knowledge, we are the first to use bootstrapped CDQNs for quantifying epistemic risk, which gives us freedom to model distributions on plausible MDPs without any structural assumptions, e.g. Gaussian distribution on parameters of Bayesian NNs or Gaussian distribution on state transitions~\citep{clements2019estimating}. An additional difference with prior work is that we use a follow the regularised leader (FTRL) algorithm to weigh the ensemble members in order to improve our uncertainty estimates.

\section{Background}\label{sec:background}

\subsection{Risk Measures: Coherence}
The idea of quantifying risk in decision making is long-studied in decision theory and has found multiple applications in finance and actuarial science.
A \textit{risk measure} maps a real-valued distribution to a real number, and quantifies the probability of occurrence of an event away from the expectation~\citep{szego2002measures}.
Some well-known risk measures are variance, Value at Risk (VaR) and Conditional Value at Risk (CVaR). \textit{Coherent} risk measures obey a set of axioms~\cite{coherent_risk}: normalisation, monotonicity, sub-additivity, homogeneity, and translation invariance.
Not all risk measures are coherent: CVaR is coherent, but variance and VaR do not satisfy respect homogeneity and subadditivity respectively~\citep{coherent_risk}.

If a coherent risk measure also satisfies comonotonic subadditivity~\citep[Axiom 4]{SONG2009459}, it can be expressed as an expectation over a distorted distribution function, for a concave \textit{distortion function} $U_\alpha:[0,1] \to [0,1]$.  Specifically (see~\citep[Theorem 2]{WANG1997173}) a random variable $Z$ with associated probability measure $P$ and cumulative distribution function $F_Z$ satisfies: 
\begin{align}
    &\mathrm{Risk}_{U_\alpha}(Z) \triangleq \int_{\mathcal{Z}} Z \dd(U_{\alpha}\circ P)\notag\\ 
    &= \int_{\mathcal{Z}} U_{\alpha}(1-F_Z(z))\dd z = \int_0^1 U_{\alpha}(t)\dd q(1-t),
\end{align}
where $(U_{\alpha}\circ P)(A) \defn U_{\alpha}[P(A)]$ for any $A\subseteq \mathcal{Z}$.
The last line is obtained from substitution of variables~\citep{wirch2001distortion}.
Here, $q$ is the quantile function, i.e. $q(1-t) = \inf\{z\geq 0|F_Z(z) \geq 1-t \} = F_Z^{-1}(1-t)$, $U(0)=0$, and $U(1)=1$.
Since in this paper we use the risk measures for decision making, we represent a coherent risk measure through its corresponding \emph{distortion function} $U_{\alpha}$.

In this paper we focus on the \textit{CVaR}~\citep{rockafellar2000optimization} risk measure. It is extensively used in risk-sensitive RL as it is coherent, applies to general $L_p$ spaces, and captures the heaviness of the tail of a distribution. It is the expectation of the worst $\alpha$-quantile of a probability distribution, with $\alpha \in [0,1]$:
\begin{align}\hspace*{-1em}
CVaR_\alpha(Z) &\triangleq \mathbb{E}[Z \, | \, Z \leq \nu_\alpha \wedge \Pr(Z \geq \nu_\alpha) = 1-\alpha].
\end{align}
For CVaR, $U_{\alpha}(t) = \min \{ \frac{t}{1-\alpha},1\}$, 
For $\alpha = 1$, CVaR reduces to the expected value, and thus risk-neutrality.

Due to generality of our methodology and the composite risk formulation, we are able to incorporate other coherent risk measures such as the Wang risk measure~\citep{Wang2002ARM}, and standard deviation~\citep{pcirillo} (Fig.~\ref{fig:risk}).

\subsection{RL: MDP and Distributional RL}\label{sec:drl}
\noindent\textbf{MDPs.} We consider problems that can be modelled by a Markov Decision Process (MDP)~\citep{sutton2018reinforcement}. An MDP is a tuple $\mdp \triangleq (\mathcal{S}, \mathcal{A}, \mathcal{R}, \mathcal{T}, \gamma)$. $\mathcal{S} \in \mathbb{R}^d$ is a state space of dimension $d$. $\mathcal{A}$ is the set of admissible actions. $\mathcal{T}$ is a transition kernel that determines the probability of successor states $s'$ given the present state $s$ and action $a$. The reward function $\mathcal{R}$ quantifies the goodness of taking action $a$ in state $s$. 
In the risk-neutral setup, the goal of the agent is to find a policy $\pi: \mathcal{S} \rightarrow \mathcal{A}$ to maximise expected value of cumulative rewards given a time horizon $T$: 
$V^\pi(s,a) = \mathbb{E}\left[\sum_{t=0}^T \gamma^{t}R(s_t, a_t)\right]$.
Here, $s_t \sim \mathcal{T}(.|s_{t-1},a_{t-1})$, $a_t = \pi(s_t)$, $s_0=s$, $a_0=a$, and the discount factor $\gamma \in (0,1)$.
\noindent\textbf{Distributional RL.}
The variable at the core of both risk-neutral and risk-sensitive RL is usually the accumulated discounted reward $Z^\pi(s,a) \triangleq \sum_{t=0}^T \gamma^{t}R(s_t, a_t)$.
$Z^\pi(s,a)$ is called the return of a policy $\pi$.
In distributional RL, the goal is to learn the return distribution $Z^{\pol}(s, a)$ obtained by following policy $\pol$ from state $x$ and action $a$ under the given MDP.


In this work, we choose to extend CDQN by~\citet{bellemare2017distributional}, as it permits richer representations of distributions, and flexibility to compute different statistics. The intuition of using this distributional framework for risk-sensitive RL is its flexibility to model multimodal and asymmetrical distributions, which is important for an accurate estimate of risk.

\section{Quantifying Composite Risk}
\label{sec:risk}
In risk-sensitive RL, we encounter two types of uncertainties: \textit{aleatory} and \textit{epistemic}.
Aleatory uncertainty is engendered by the stochasticity of the MDP model $\mdp$ and the policy $\pol$. 
Epistemic uncertainty exists due to the fact that the MDP model $\mdp$ is unknown. In the Bayesian setting, this is represented as a belief distribution $\beta$ over a set of plausible MDPs $\Theta$. Hence, risk measures can also be defined with respect to the MDP distribution.
Consequently, as an agent learns more about the underlying MDP, the epistemic risk vanishes.
The aleatory risk is inherent to the MDP  $\mdp$ and policy $\pol$, and thus persists even after correctly estimating the model $\mdp$. Let us now define risk measures for aleatory and epistemic uncertainties, and then combine them into a composite risk measure.

\textbf{Aleatory Risk.}
Given a coherent risk measure with distortion function $U^A_{\alpha}$, the aleatory risk is quantified as the deviation of total risk of individual models from the risk of the average model.
\begin{align*}
    A(U^A_{\alpha}, \beta) &\triangleq \int_{\Theta} \int_{\mathcal{Z}} Z \dd(U^A_{\alpha} \circ \Pr)(Z|\theta) \dd\beta(\theta) \\&- \int_{\Theta} \int_{\mathcal{Z}} \hat{Z} \dd(U^A_{\alpha} \circ \Pr)(\hat{Z}) 
\end{align*}
\ifsinglecol
\begin{align*}
A(U^A_{\alpha}, \beta) &\triangleq \mathbb{E}_{\theta \sim \beta}[\sup_{Q\in Q^{\theta}_{\alpha}}\mathbb{E}[Z_\theta] - \sup_{Q\in Q^{\hat{\theta}}_{\alpha}}\mathbb{E}[Z_{\hat{\theta}}]] = \mathbb{E}_{\theta \sim \beta}[\sup_{Q\in Q^{\theta}_{\alpha}}\mathbb{E}[Z_\theta]] - \sup_{Q\in Q^{\hat{\theta}}_{\alpha}}\int_{\Theta}\int_{\mathcal{Z}}z \dd Q(z|\theta)\dd\beta(\theta).
\end{align*}
\else
\fi
Here, $\Pr(\hat{Z}) = \int_{\Theta} \Pr(Z|\theta)\dd\beta(\theta)$, i.e. the return distribution of the average model. The centered definition of aleatory risk is necessary to show that additive risk is a special case of composite risk.

\textbf{Epistemic Risk.}
Given a coherent risk measure with distortion function $U^E_{\alpha}$, the epistemic risk quantifies the uncertainty invoked by not knowing the true model. Thus, the risk can be computed over any statistics of the models, such as expectation.
\begin{align*}
    E(U^E_{\alpha}, \beta) &\triangleq \int_{\Theta} \int_{\mathcal{Z}} Z \dd\Pr(Z|\theta) \dd(U^E_{\alpha}\circ\beta)(\theta)
\end{align*}
\ifsinglecol
\begin{align*}
E(U^E_{\alpha}, \beta) &\triangleq \sup_{\beta' \in \Beta_{\alpha}} \mathbb{E}_{\theta \sim \beta'}[\mathbb{E}_{Z\sim \Pr(.|\theta)}[Z]].
\end{align*}
\else
\fi

\textbf{Composite Risk under Model and Inherent Uncertainty.}
In typical risk-sensitive RL settings, the true MDP model is both unknown and inherently stochastic. Thus, the overall uncertainty is a composition of aleatory and epistemic uncertainties. For that reason, quantify it using what we call the \textit{composite risk}.
\begin{definition}[Composite Risk]\label{def:composite}
For two coherent risk measures with distortion functions $U^A_{\alpha_1}$ and $U^E_{\alpha_2}$, belief distribution $\beta$ on model parameters $\theta \in \Theta$, and a random variable $Z \in \mathcal{Z}$, the composite risk of epistemic and aleatory uncertainties is defined as
\begin{align}\label{eq:composite}
    &F^C(U^A_{\alpha_1}, U^E_{\alpha_2}, \beta) \triangleq \mathrm{Risk}_{\UE}(\mathrm{Risk}_{\UA}(Z|\theta)|\beta)\notag\\
    &= \int_{\Theta} \int_{\mathcal{Z}} Z \dd(U^A_{\alpha_1} \circ \Pr)(Z|\theta) \dd(U^E_{\alpha_2} \circ \beta)(\theta)\notag\\
    &= \int_0^1 \int_0^1 U^E_{\alpha_2}(v) U^A_{\alpha_1}(u)\dd q_{Z|\theta}(1-u) \dd q_{\beta}(1-v) 
\end{align}
\end{definition}
Here, $q_{Z|\theta}$ and $q_{\beta}$ are quantile functions of $Z$ conditioned on $\theta$ and that of $\theta$ respectively. For brevity, we also denote $F^C(U^A_{\alpha_1}, U^E_{\alpha_2}, \beta)$ as $\mathrm{Risk}_{\UE}\circ\mathrm{Risk}_{\UA}$ (e.g. $\mathrm{CVaR} \circ \mathrm{CVaR}$), whenever it is clear from the context.
\begin{theorem}[Coherence]\label{thm:coherence}
	If $U^A_{\alpha_1}$ and $U^E_{\alpha_2}$ are distortion functions for two coherent risk measures, the composite risk measure $F^C(U^A_{\alpha_1}, U^E_{\alpha_2}, \beta)$ is also coherent.
\end{theorem}
The proof of Theorem~\ref{thm:coherence} is available in Appendix~\ref{sec:proof}. The generic nature of our composite risk definition allows us to use different risk measures compatible with epistemic and aleatory risks. This is demonstrated in experiments (Figure~\ref{fig:risk}) using different combinations of CVaR, Wang risk, and standard deviation for quantifying epistemic and aleatory uncertainties. This flexibility was absent in previous risk-sensitive RL literature~\citep{eriksson2019epistemic,depeweg2018decomposition,clements2019estimating}. 

\textbf{Comparison with Additive Risk Formulations.} \cite{clements2019estimating, depeweg2018decomposition} use a weighted sum of epistemic and aleatory variances as their risk measure. This formulation has mainly two problems. First, variance is not a coherent risk measure as it does not follow the homogeneity and subadditivity properties, as shown in~\citep{pcirillo}. Secondly, we show that even if we replace the variance with a coherent risk measure, the additive formulation is equivalent to considering $U^E_{\alpha}$ as an identity function. Thus, it is less sensitive to the effect of epistemic uncertainty than composite risk. More formally:

\begin{theorem}\label{thm:comp_geq_add}
	We are given two sources of aleatory and epistemic uncertainties $\xi_1$ and $\xi_2$. If $U^A_{\alpha_1}$ and $U^E_{\alpha_2}$ are distortion measures for two coherent risk measures quantifying aleatory and epistemic risks respectively, then, i) $F^A(U^A_{\alpha_1}, \beta) = F^C(U^A_{\alpha_1}, I, \beta)$, where $I$ is the identity function, and ii) $F^C(U^A_{\alpha_1}, U^E_{\alpha_2}, \beta) \geq F^A(U^A_{\alpha_1}, \beta)$, if $\alpha_2 \neq 1$.
\end{theorem}

\begin{figure}[t!]
	\centering
	\includegraphics[width=0.4\textwidth]{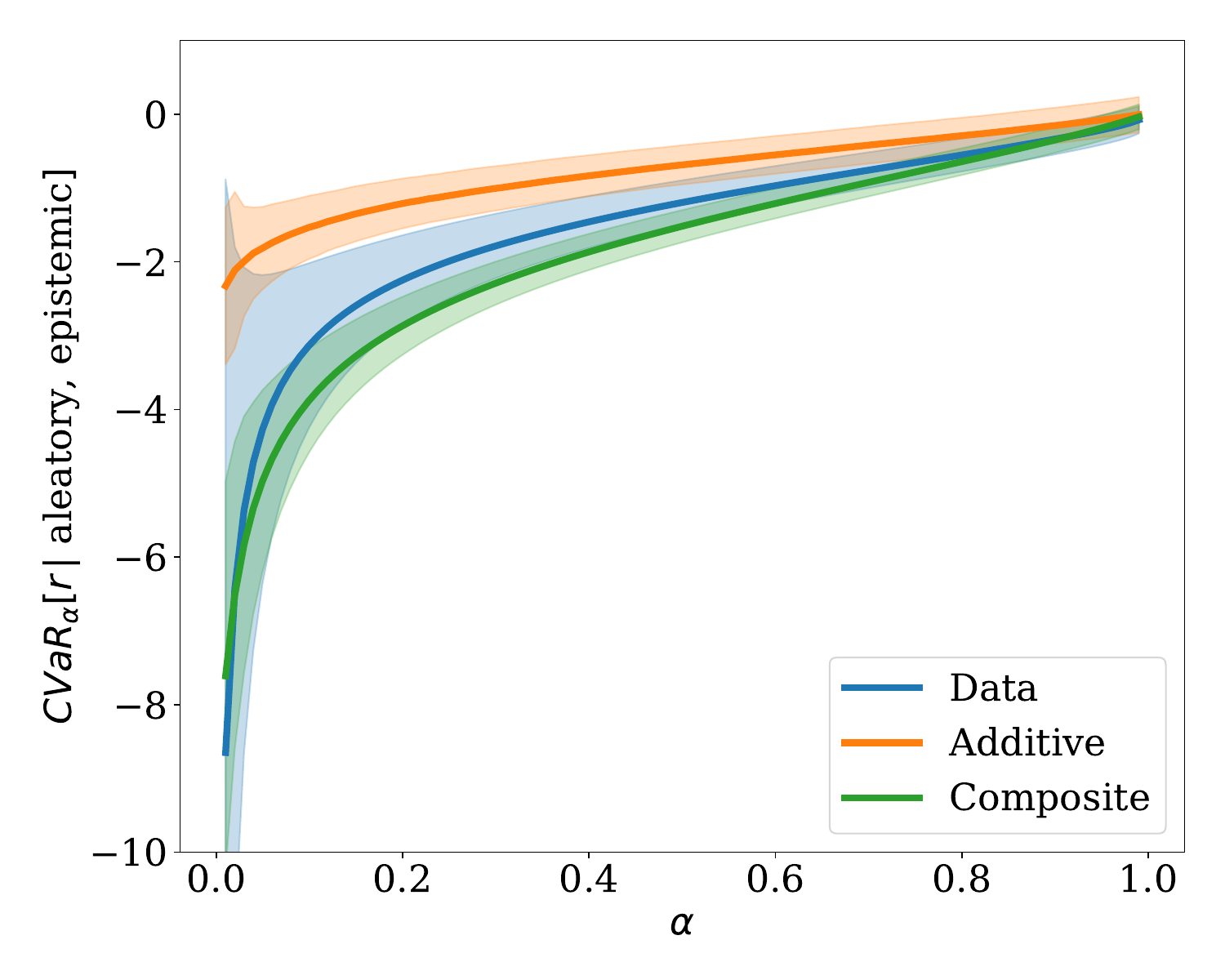}
	\caption{Estimation of total $CVaR_{\alpha}$ from a mixture of 100 Gaussians sampled from a posterior distribution. Total $CVaR_{\alpha}[Data]$ is based on the marginal distribution of $r$ as in Example~\ref{ex:gauss-mix-cvar}. We compare this with composite and additive estimates and illustrate results over $100$ runs. Here, lower value of CVaR indicates higher mass on the left tail of the distribution and higher risk of obtaining low returns.}\label{fig:gauss-mix-cvar}
\end{figure}
\begin{example}[A Reductive Empirical Evaluation of Composite and Additive Risks]\label{ex:gauss-mix-cvar}
    We consider a mixture of $100$ Gaussians: $p(r) = \sum_{i=1}^{100} \phi_i\mathcal{N}(\mu_i, \sigma_i^2)$, where $\Phi \sim Dir([0.5]^{100}), \mu \sim \mathcal{N}(0, 1)$, and $\sigma^2 \sim \Gamma^{-1}(2, 0, 1)$.
	We compute $CVaR_{\alpha}[r]$ using the data generated from this mixture over 100 runs. We further estimate composite risk with $U_E, U_A = CVaR_{\alpha}$ and additive risk with $U_A= CVaR_{\alpha}$. The results illustrated in Figure~\ref{fig:gauss-mix-cvar} show that the additive CVaR risk strictly underestimates the total CVaR risk computed from the data, whereas the composite risk is closer to the one computed from data. Specifically, for lower values of $\alpha$ (specifically, $\alpha \leq 0.5$), i.e. towards the extreme end of the left tail where events occur with low probability, the additive CVaR risk deviates significantly from data whereas the composite measure yields closer estimation. Such values of $\alpha$'s are typically interesting for risk-sensitive applications.
\end{example}

This means that for given sources of aleatory and epistemic uncertainties the additive risk which only considers expectation over epistemic uncertainty will always underestimate the composite effect of epistemic risk. 
Thus, we observe that additive risk leads to worse risk-sensitive performance than composite risk in RL problems (Table~\ref{tab:highway10} and Figure~\ref{fig:highway}). 

\section{Algorithm: SENTINEL-K}\label{sec:algo}
Now, we outline the algorithmic details of SENTINEL-K that estimates composite risk over returns using an ensemble of $K$ distributional RL estimators, namely CDQN, in tandem with an adaptation of FTRL for estimator selection, and leverage the estimates for decision making.

\textbf{Sketch of the Algorithm.}
Pseudocode of SENTINEL-K with composite risk is given in Algorithm~\ref{alg:composite}.
 It has two main blocks: obtaining $K$ estimates of return distribution with distributional RL framework (Lines~\ref{line:block1_1}-~\ref{line:block1_2}), and using them to compute composite risk for each action (Lines~\ref{line:block2_1}-~\ref{line:block2_2}). Finally, following the mechanism of Q-learning~\citep{watkins1992q}, it chooses the action with maximal composite risk in the decision making step (Line~\ref{line:action}).
 
 In the first block (Lines~\ref{line:block1_1}-~\ref{line:block1_2}), we specifically use an ensemble of $K$ CDQNs. Each CDQN uses target and value networks for estimating the return distribution. We set a schedule for updating the target networks $\Gamma_1$ and a more frequent one ($\Gamma_1 \cup \Gamma_2$) for the value networks (Section~\ref{sec:ensemble}). 
 
 The second block (Lines~\ref{line:block2_1}-~\ref{line:block2_2}) is used for decision-making and iterated at every time step. It adapts the FTRL algorithm (Section~\ref{sec:ftrl}) for aggregating the $K$ estimated return distributions and to compose aleatory risk $Q^A_i(s_t, a)$ of each of the estimators to provide a final estimate of the composite risk $Q^C(s_t, a)$ for each action, and then selecting the action with highest $Q^C(s_t, a)$.

\subsection{Ensembling and Bootstrapping $K$-Estimators}\label{sec:ensemble}
The ensemble of SENTINEL-K consists of $K$ distribution estimators. Each estimator gets its own dataset $\lbrace D_i\rbrace_{i=1}^K \subseteq \mathcal{D}$, value network $\lbrace \theta_{i}\rbrace_{i=1}^K$ and target network $\lbrace \theta_{i}^-\rbrace_{i=1}^K$. 
The $K$ datasets are created from the original dataset $\mathcal{D}$ by \textit{data masking} (Line~\ref{line:datamask}). 
For each transition $s_t, a_t, r_t, s_{t+1}$, a fixed weight vector $\mathbf{u}_t \in [0, 1]^K$ is generated such that $u^j_t \sim Ber(\frac{1}{3})$. 
Thus, on an average, each estimator $i$ has access to $\frac{1}{3}$ of the dataset. Details about data masking are in Appendix~\ref{sec:datamask}.

After preparing the datasets for the estimators, the target and value networks of the CDQN have to be updated and optimised.
For $i$-th estimator, it begins with sampling mini batches of data $\tau$ from the respective dataset $D_i$ (Line~\ref{line:minibatch}).
Then, this dataset is used to compute the composite risk for all actions $a \in \mathcal{A}$ and to obtain $a^{*}$ (Lines~\ref{line:f_estimate}-~\ref{line:astar}). 
Obtaining the composite risk first involves estimating the aleatory risk with $Q_i^A(s_t, a)=\int_{\mathcal{Z}} Z \dd(\UA \circ \Pr)(Z|\theta_i)$ for a particular estimator $i$. This quantity can be attained by considering each of the estimators separately, however, as we turn to compute the epistemic risk the estimators jointly contribute to this risk.
Then, we compose the aleatory risk of all the estimators to compute $Q^C(s_t, a) = \mathrm{Risk}_{\UE}(\{Q_i^A(s_t, a)\}_{i=1}^K)$. 
Here, $\mathrm{Risk}_{\UE}$ is the risk measure corresponding to the distortion $\UE$. Finally, the optimal action $a^{*} = \underset{a}{\arg\max} \, Q^C(s_t, a)$, and the risk estimates $Q^C(s_t, a)$ are used to update the value and network parameters $\lbrace \theta_{i}\rbrace_{i=1}^K$ and $\lbrace \theta^-_{i}\rbrace_{i=1}^K$ (Lines~\ref{line:valuenet}-~\ref{line:targetnet}) by minimising the cross-entropy loss of the current parameters and the projected Bellman update as described in~\citep{bellemare2017distributional}.

Ensembling estimators have been shown to outperform individual estimators as seen in~\citep{wiering2008ensemble, fausser2015neural, osband2016deep, pacchiano2020optimism}.
Further, incorporating multiple estimators introduces uncertainty over the estimators. 
Because of having separate data sets, each of the estimators learn different parts of the MDP.
Thus, uncertainty over estimators acts as a quantifier of the model uncertainty. In Section~\ref{sec:experiments}, we show that this ensemble-based approach leads SENTINEL-K to achieving superior performance.

\subsection{Weighing Estimates with FTRL}\label{sec:ftrl}
Now, the question is to adaptively and accurately aggregate the $K$ estimated return distributions.
\cite{pacchiano2020optimism} shows that adaptive model selection can boost performance in comparison to model averaging.
The rationale for this can be given by seeing that some estimators might be overly optimistic or pessimistic.
By weighing these less, you can effectively have a more robust ensemble. Further discussion of this issue is given in Appendix~\ref{sec:appendix_ftrl}.

We adapt the Follow The Regularised Leader (FTRL) algorithm~\citep{cesa2006prediction} studied in bandits and online learning for adaptively weighing the estimators. FTRL puts exponentially more weight on an estimator depending on its accuracy of estimating the return distribution. Since we do not know the `true' return distribution, we use the KL-divergence from the posterior of a single estimator $i$, $\Pr(Z \, | \, \theta_i)$, to the posterior marginalised over $\bel(\theta)$, i.e. $l(\theta_i, \beta) \triangleq
D_{\mathrm{KL}}\Big(\Pr(\hat{Z})\, || \, \Pr(Z \, | \, \theta_i) \Big)$, 
as proxy of estimation loss of estimator $i$.
FTRL selects estimator $i$ with weight
\begin{equation}\label{eq:weights}
w_i = \dfrac{e^{\lambda l(\theta_i, \beta)}}{ \sum_j e^{\lambda l(\theta_j, \beta)}}, \quad\lambda\in [0,\infty).
\end{equation}
Using FTRL weights for aggregating the $K$ return distributions is analogous to using an exponentially weighted average forecaster~\citep{cesa2006prediction} on the $K$ learners to create a final estimate of the return distribution and corresponding composite risk. This leads to a better aggregation of individual estimates than equally weighted average or a greedy selection of the best estimate~\citep[Theorem 2.2]{cesa2006prediction}.
%
Having computed the weights $\mathbf{w}$ (Line~\ref{line:ftrl}), we compute the weighted composite risk measure by first computing the aleatory risk of  each of the estimators, 
$Q^A_i(s_t, a) = \int_{\mathcal{Z}} Z \dd(\UA \circ \Pr)(Z|\theta_i)$ (Line~\ref{line:aleatory}), and then the composite risk is computed by $Q^C(s_t, a) = \mathrm{Risk}_{\UE}(\{w_i Q_i^A(s_t, a)\}_{i=1}^K)$ (Line~\ref{line:composite}).
Here, $\lambda \in [0, \infty)$ is a regularising parameter that determines to what extent estimators far away from the marginal estimator should be penalised.
If $\lambda \rightarrow 0$, we obtain standard model averaging. If $\lambda \rightarrow \infty$, it reduces to greedy selection. We experimentally show that performing FTRL with a reasonable $\lambda$ value, namely 1, leads to better performance.

\textbf{Action Selection.} The algorithm always selects the action with the high composite risk $Q^C$. Its behaviour depends on the choice of risk measures or distortion utility functions $\UA$ and $\UE$. SENTINEL-K reduces to a risk-neutral algorithm if we choose both $\UA, \UE$ as identity functions, and to additive risk-sensitive algorithm if we choose $\UE$ as identity. Designing it to accommodate composite risk provides us the flexibility to be risk-sensitive, risk-neutral, and treating epistemic and aleatory risk with different metrics. 

\begin{algorithm}[t!]
	\caption{SENTINEL-K with Composite Risk}
	\label{alg:composite}
	\begin{algorithmic}[1]
		\STATE{\textbf{Input:} } Initial state $s_0$, action set $\mathcal{A}$, distortion measures $\UA, \UE$, hyperparameter $\lambda$, target networks
		$[\theta_1^{-}, ..., \theta_K^{-}]$, value networks
		$[\theta_1, ..., \theta_K]$, update schedule $\Gamma_1, \Gamma_2$.
		\FOR{$t=1,2,\ldots$}
		\STATE //* Update $K$-value and target networks for estimating return distributions *//
		\FOR{$t' \in \Gamma_1 \cup \Gamma_2$ }\label{line:block1_1}
		\STATE  Generate $\{D_1, ..., D_K\} \gets \mathrm{DataMask}(\mathcal{D}^{t'})$\label{line:datamask}
		\FOR{$i = 1, \ldots, K$}
		\STATE Sample mini batch $\tau \sim D_i$\label{line:minibatch}
		\STATE Estimate \eqref{eq:composite} $F^C(Z(s_t,a)|\UA, \UE, \beta)$ using $\tau$ and $K$-target networks $\lbrace \theta_{i}^-\rbrace_{i=1}^K$. \label{line:f_estimate}
		\STATE Get $a^{*} = \argmax_a F^C(Z(s_t,a)|\UA, \UE, \beta)$\label{line:astar}
		\STATE  Update value network $\theta_i$ using $\tau, a^{*}$\label{line:valuenet}
		\STATE  Update target network $\theta^{-}_i$ using $\tau, a^{*}$ if $t' \in \Gamma_1$\label{line:targetnet}
		\ENDFOR
		\ENDFOR \label{line:block1_2}
		\STATE //* Estimate the composite risk of each action using the estimated return distributions *//
		\FOR{$a \in \mathcal{A}$}\label{line:block2_1}
		\STATE Compute weights $\mathbf{w} = w_1, ..., w_K$ from Eq.~\ref{eq:weights}.\label{line:ftrl}
		\FOR{$i$ \textbf{in} $K$}
		\STATE Compute aleatory risks $Q^A_i(s_t, a)$ from $\int_{\mathcal{Z}} Z \dd(\UA \circ \Pr)(Z|\theta_i)$ \label{line:aleatory}
		\ENDFOR
		\STATE Compute composite risk over weighted aleatory estimates $Q^C(s_t, a) = \mathrm{Risk}_{\UE}(\{w_i Q_i^A(s_t, a)\}_{i=1}^K)$\label{line:composite}
		\ENDFOR \label{line:block2_2}
		\STATE //* Action selection *//
		\STATE{Take action $a_t = \argmax_a Q^C(s_t, a)$}\label{line:action}
		\STATE 	Observe $s_{t}$ and update the dataset $\mathcal{D}^{t} \gets \mathcal{D}^{t-1} \cup \{s_t, a_{t-1}, s_{t-1}, r_{t-1}\}$\label{line:update}
		\ENDFOR 
	\end{algorithmic}
\end{algorithm}



\section{Experimental Evaluation}\label{sec:experiments}
We test the risk-sensitive performance of SENTINEL-K with composite CVaR risk in two environments with continuous state spaces. We also display the flexibility of our composite risk formulation by evaluating heterogeneous risks with SENTINEL-K.\footnote{\noindent Ablation studies for risk-neutral SENTINEL are in Appendix.}
Settings for each of these experiments and results are elaborated in corresponding subsections.
In all the experiments, we use $4$ CDQNs in the ensemble and call it SENTINEL-4. We justify this choice of $K=4$ in Appendix~\ref{sec:ensemble_size}. For each experiment, we report the mean and standard error of the mean over 20 runs for $10^5$ steps.
\begin{table*}[t!]
	\centering
	\caption{Performance of risk-neutral (VDQN, CDQN, SENTINEL-K), aleatory risk-sensitive VDQN-CVaR, UA-DQN and risk-sensitive (SENTINEL-4 with additive and composite CVaRs) for highway-v1 with 10 vehicles. Results are reported over 20 runs. SENTINEL-4 with composite CVaR performs better.}\label{tab:highway10}
	\resizebox{0.8\textwidth}{!}{	\begin{tabular}{c|c|c|c}
		Agent & Value $\pm \sigma$ & Aleatory metric $\pm \sigma$ & $\#$ crashes $\pm \sigma$ \\
		\hline
		VDQN$_{RN}$~\cite{tang2018exploration}& $23.30\pm0.36$ & $14.29\pm0.80$ & $1252.33\pm170.35$ \\
		CDQN$_{RN}$~\cite{bellemare2017distributional}& $25.96\pm0.51$ & $19.50\pm1.44$ & $839.53\pm150.20$\\
		SENTINEL-4$_{RN}$& $26.56\pm0.32$ & $20.88\pm 1.25$& $617.11\pm 100.15$\\
		VDQN-CVaR$_A$~\cite{tang2018exploration} & $24.39\pm0.50$ & $16.64\pm1.25$ & $871.33\pm171.23$\\
		UA-DQN$_{E+A}$~\cite{clements2019estimating}& $24.46\pm0.29$ & $16.9\pm0.44$ & $1060.65\pm13.94$\\
		SENTINEL-4$_{E+A}$& $26.82\pm0.42$ & $21.54\pm1.40$ & $645.55\pm127.59$\\
		SENTINEL-4$_{E\circ A}$& $\mathbf{27.43\pm0.13}$ & $\mathbf{24.16\pm0.54}$ & $\mathbf{341.18\pm43.86}$\\
	\end{tabular}
}
\end{table*}

\textbf{Risk-sensitive Performance.} In order to demonstrate performance in a larger domain, we opt to evaluate SENTINEL-4 in the \textit{highway}~\citep{highway-env} environment.
Highway is an environment developed to test RL for autonomous driving. 
We use a version of the \textit{highway-v1} domain with five lanes, and ten vehicles in addition to the ego vehicle.
In this environment, the episode is terminated if any of the vehicles crash or if the time elapsed is greater than $40$ time steps. 
The reward function is a combination of multiple factors, including staying in the right lane, the ego vehicle speed, and the speed of the other vehicles.

We test the risk-neutral CDQN and VDQN algorithms, an aleatory risk-sensitive VDQN and the total variance decomposition algorithm UA-DQN along with SENTINEL-4 with both additive and composite CVaRs.
The typical performance metric for this scenario is the expected discounted return $\mathbb{E}_\mdp^{\pol}[R]$. 
In order to test the risk-sensitive performance, we use two metrics.
In order to measure aleatory risk $\UA[R \, | \, \pol, \mdp]$, we use CVaR as $\UA$ with threshold $\alpha = 0.25$.
The CVaR metric is a statistic of the left-tail of the return distribution and higher values would mean better performance in the $25\%$ worst-cases of performance. 
Finally, as a proxy for the epistemic risk, we use the number of crashes (lower is better).

\begin{figure*}[t!]
\begin{minipage}[c]{0.32\textwidth}
	\centering
	\includegraphics[width=1\textwidth]{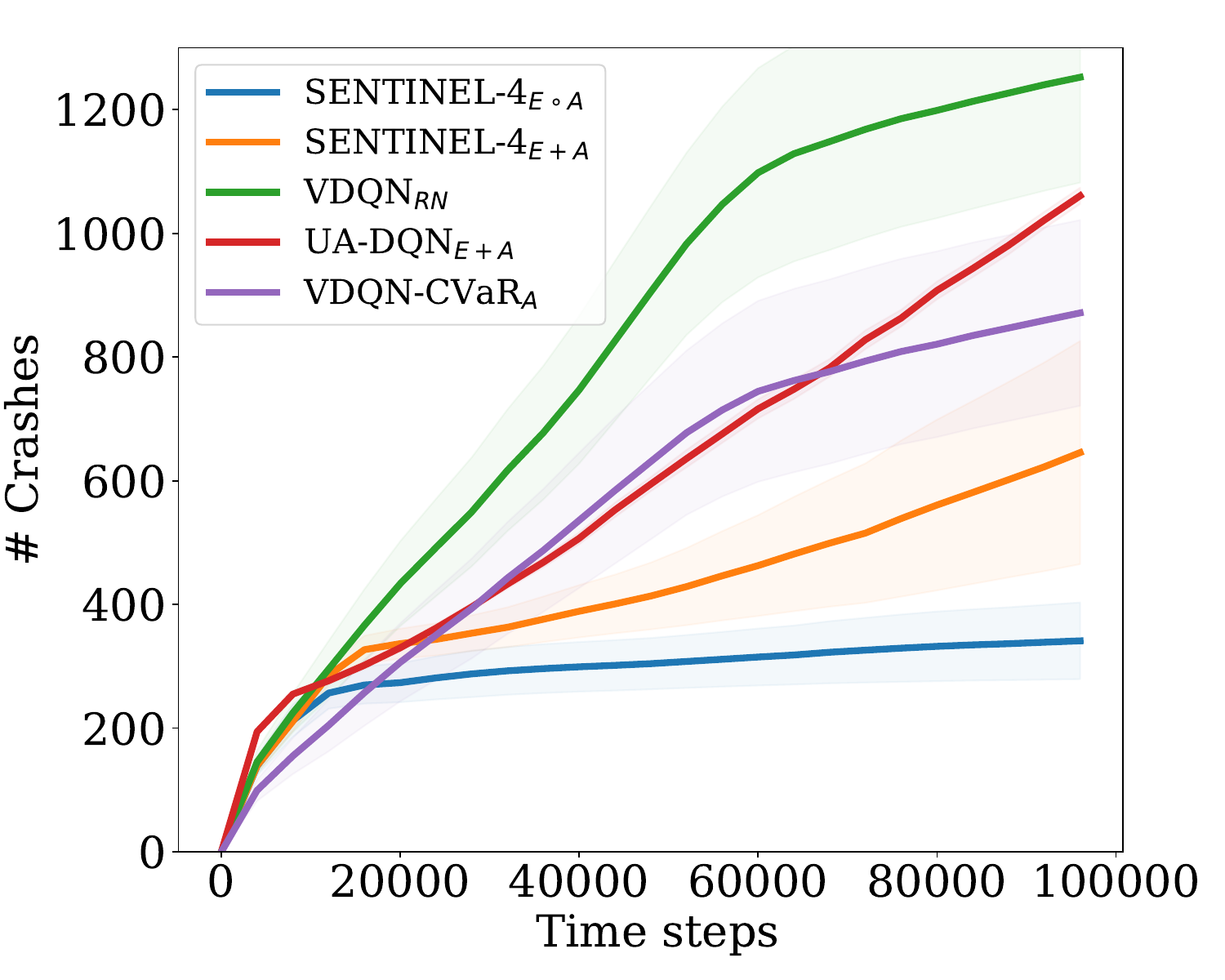}
	\caption{The total number of crashes in highway environment with $10$ vehicles over $20$ runs and horizon $10^6$. Fewer \#crashes indicate better risk-sensitive performance.} 
	\label{fig:highway}
\end{minipage}\hfill
\begin{minipage}[c]{0.32\textwidth}
	\centering
	\includegraphics[width=\textwidth]{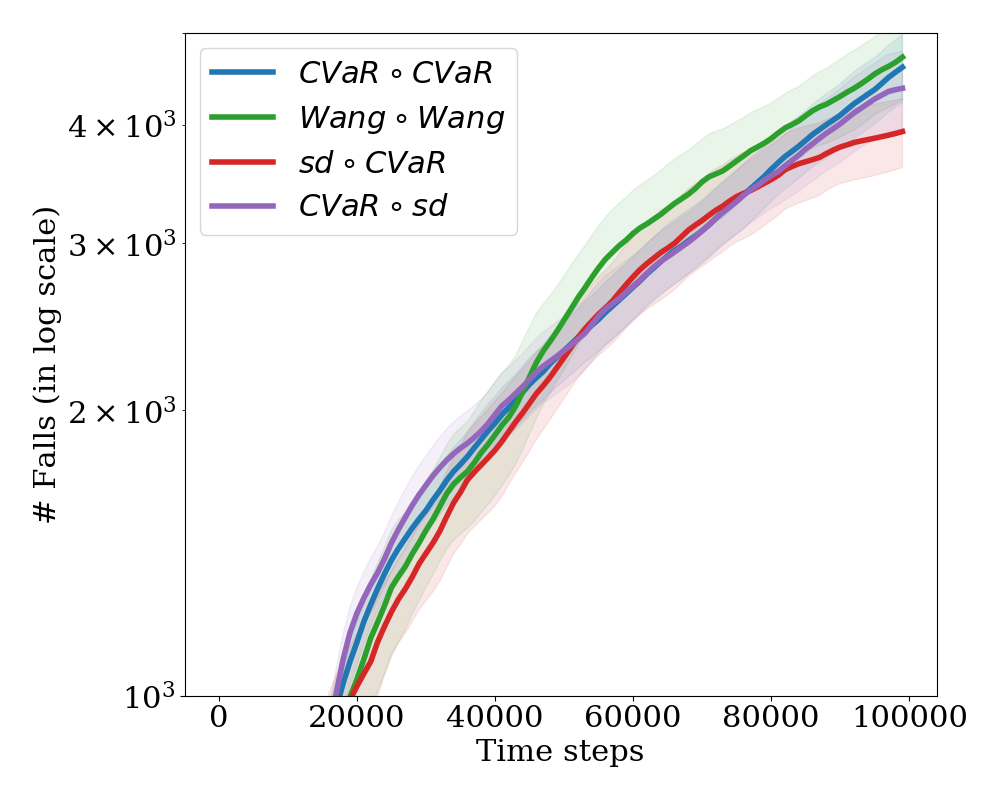}%
	\caption{Performance and convergence of SENTINEL-4 using different risk measures. 
	We show the number of falls in the \textit{CartPole} environment over $20$ runs with different initialisation.}\label{fig:risk}
\end{minipage}\hfill
\begin{minipage}[c]{0.32\textwidth}
	\centering
	\includegraphics[width=\textwidth]{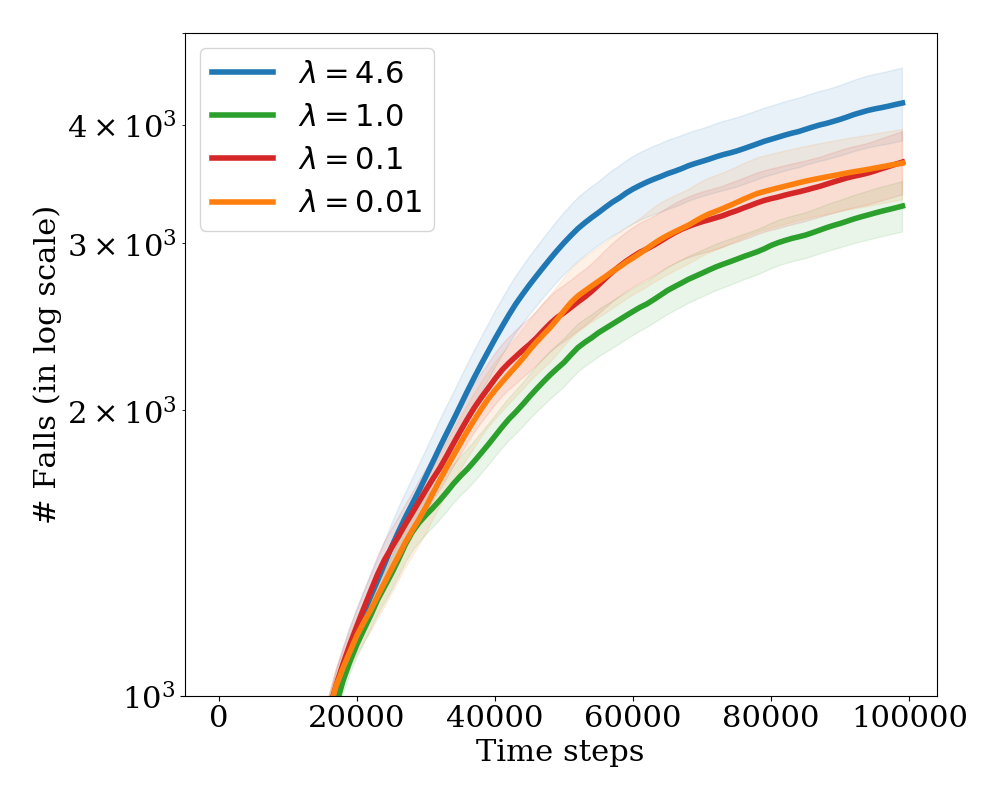}%
	\caption{Performance and convergence of SENTINEL-4 (risk-neutral) for different values of $\lambda$. We show the number of falls in \textit{CartPole} environment over $20$ runs with different initialisation.}\label{fig:cartpole}
\end{minipage}
\end{figure*}

Experimental results are illustrated in Table~\ref{tab:highway10} and Figure~\ref{fig:highway}.
From Table~\ref{tab:highway10}, we observe that our algorithm with composite risk achieves a higher value, higher estimate of aleatory risk, and less number of crashes.
Thus, SENTINEL-4 with composite CVaR outperforms the competing algorithms in terms of all three metrics. The simultaneous improvement in both the value function and \#crashes is due to the fact that \textit{highway} is designed to have a reward function that penalises unsafe driving.
Additionally, we observe that the variance of performance metrics over 20 runs is the least for our algorithm with composite CVaR measure. 
This shows the stability of our algorithm which is another demonstration of good risk-sensitive performance.
Figure~\ref{fig:highway} resonates with these observations in terms of the total number of crashes.

\textbf{Heterogeneous Risk Measures.}
In order to demonstrate the flexibility of the composite risk framework estimated with SENTINEL, we investigate performance using heterogeneous coherent risk measures, that composes different coherent risk measures for aleatory and epistemic risk. The chosen risk measures are aleatory and epistemic CVaR, aleatory and epistemic Wang risk, aleatory CVaR with epistemic standard deviation, and aleatory standard deviation with epistemic CVaR. Note that any combination of coherent risk measures is possible. 
We evaluate SENTINEL-4 in the \textit{CartPole-v0} environment~\citep{openAI16}. 
This environment is a popular test-bed for continuous state-space RL tasks. 
In the environment, a reward of $1$ is attained for every time step the pole is kept upright. 
If the pole falls to either of the sides or if the number of time steps reaches $200$, the episode is terminated. 
This means that the undiscounted return attained per episode is in $[0, 200]$. Thus, we choose $V_{min} = 0, V_{max} = \frac{1-\gamma^{200}}{1-\gamma}$ as the histogram support of CDQN. The results are shown in Figure~\ref{fig:risk}, which demonstrates than SENTINEL-4 performs flexibly and comparably for these composite risks.

\textbf{FTRL vs. Average vs. Greedy.} We choose $[0.01, 0.1, 1.0, \ln 100]$ as the different values of the regularising hyperparameter $\lambda$ and test the performance of SENTINEL-4 for \textit{CartPole-v0}. As $\lambda \rightarrow 0$, we perform standard model averaging which is sensitive to outliers.
As $\lambda \rightarrow \infty$, model selection gets greedily biased towards the best average estimator while not providing other estimators a chance to improve. 
A sound value of $\lambda$ would be one that excludes outlier estimators while still involves most of the other estimators. 
Figure~\ref{fig:cartpole} shows performance in terms of cumulative $\#$ Falls (lower is better) for the $\lambda$ values with $CVaR_{0.25}\circ CVaR_{0.25}$.
We observe that FTRL with reasonable $\lambda=1.0$ shows better performance, i.e. less number of falls, than the ones with large $\lambda=4.6$ and small $\lambda$'s $0.01$ and $0.1$.
We also observe that for $\lambda=1$ the variance of $\#$Falls is significantly less than that of other values and thus, stability of performance.

\textbf{Summary of Results.} 
Fig.~\ref{fig:highway} shows the risk-sensitive performance of VDQN, CDQN, aleatory CVaR, total variance decomposition UA-DQN and SENTINEL-4 additive and composite CVaR risks on a large continuous state environment. SENTINEL-4 with composite risk outperforms competing algorithms in terms of the achieved value function and estimated aleatory risk. It causes the least number of crashes than competing algorithms.
Fig.~\ref{fig:risk} demonstrates the ability to chose any coherent risk measure for SENTINEL-K, including different risk measures for both epistemic and aleatory risk.
Fig.~\ref{fig:cartpole} shows that selecting $\lambda$ is important in bootstrapped RL, and tuning it yields better performance over model averaging ($\lambda \rightarrow 0$) and greedy selection ($\lambda \rightarrow \infty$).
We defer the results on the choice of $K$ in ensemble, convergence in return distribution, and improved efficiency in estimating multi-modal return distributions, to Appendix. 



\section{Discussion}\label{sec:discussion}
In this paper, we study the problem of risk-sensitive RL. We propose two main contributions. 
The first is the \textit{composite risk} formulation that quantifies the holistic effect of aleatory and epistemic risk involved in learning. With a reductive experiment, we show that composite risk estimates the total risk involved in a problem more accurately than existing additive formulations.
The second one is \textit{SENTINEL-K} which ensembles $K$ distributional RL estimators, namely CDQNs, to provide an accurate estimate of the return distribution.  
We adopt FTRL from bandit literature as a means of model selection. 
FTRL weighs each estimator adaptively and leads to better experimental performance than greedy selection and model averaging.
Experiments show that SENTINEL-K achieves superior risk-sensitive performance while used with composite CVaR estimate, and can operate on composition of different risks unlike existing works.

Motivated by the experimental success, we aim to investigate theoretical properties of FTRL-driven bootstrapped distributional RL with and without composite risk estimates. 




\begin{acknowledgements} 
We would like to thank Dapeng Liu for fruitful discussions in the beginning of the project, further, this work was partially supported by the Wallenberg AI, Autonomous Systems and Software Program (WASP) funded by the Knut and Alice Wallenberg Foundation and the computations were enabled by resources provided by the Swedish National Infrastructure for Computing (SNIC) at C3SE partially funded by the Swedish Research Council through grant agreement no. 2018-05973.
\end{acknowledgements}

\bibliography{references}

\appendix
\onecolumn
\section{Coherent Risk Measures}
In the following section we expand on details that did not make it into the main paper.
\subsection{Formal Definitions}

\begin{definition}[Coherent Risk Measure] A coherent risk measure is a mapping $U: \Delta(\mathcal{X}) \rightarrow \Reals$ from a set of distributions on $\mathcal{X}$ to the real numbers,  satisfying four axioms:
\begin{axiom}[Monotonicity]
    If $X\leq Y$ almost surely, $U(X) \leq U(Y)$.
\end{axiom}
\begin{axiom}[Positive homogeneity]
    For any $c \geq 0$, $U(cX) = c U(X)$.
\end{axiom}
\begin{axiom}[Translation invariance]
    For any constant $a \in \Reals$, $U(X+a) = U(X) + a$.
\end{axiom}
\begin{axiom}[Subadditivity]
    For $X, Y \in \mathcal{X}$, $U(X+Y) \leq U(X) + U(Y)$.
\end{axiom}
\end{definition}

\begin{definition}[Conditional Value-at-Risk]
For a random variable $Z$ quantifying risk and $\alpha \in [0,1]$,
\begin{align}
CVaR_\alpha(Z) &\triangleq \mathbb{E}[Z \, | \, Z \leq \nu_\alpha \wedge \Pr(Z \geq \nu_\alpha) = 1-\alpha]
\end{align}
\end{definition}

\begin{definition}[Entropic Value-at-Risk]
For a random variable $Z$ quantifying risk and $\alpha \in [0,1]$,
\begin{equation}
    EVaR_{\alpha}(Z) = \underset{\lambda>0}{\inf}\Big\{\lambda^{-1}\ln\Big(\frac{M_Z(\lambda)}{1-\alpha}\Big)\Big\}
\end{equation}
Here, $M_Z(\lambda) \triangleq \mathbb{E}[\exp(\lambda Z)]$ is the moment generating function (MGF) of $Z$ for any $\lambda \in \Reals$. For $\alpha=0$, EVaR reduces to entropic risk measure or exponential utility based risk measure.
\end{definition}

\begin{definition}[Wang risk measure]
    For a random variable $Z$ quantifying risk and $\alpha \in [0,1]$,
\begin{equation}
WR_{\alpha}(Z) = \Phi[\Phi^{-1}(F(Z))-\Phi^{-1}(\alpha)].
\end{equation}
Here, $F(Z)$ is the Cumulative Distribution Function (CDF) of $Z$. $\Phi$ and $\Phi^{-1}$ are the standard normal CDF and inverse normal CDF respectively. For $\alpha\leq 0.5$, this induces risk aversion and for $\alpha\geq 0.5$, it causes risk attraction.
\end{definition}

\subsection{Our Approach of Computing Risk over Return Distributions}
The aforementioned coherent risk measures can also be written as an expectation over a distorted cumulative distribution function for a given distortion function $g$. Here, $g:[0,1] \to [0,1]$, $g(0)=0$, and $g(1)=1$. When combined with a probability measure $P$ the distortion function defines a new function on events:
\[
  (g \circ P)(A) \defn g[P(A)].
\]
The distortion function allows us to treat different samples with different risk-sensitive weights unlike standard expectation where $g(t) = t$. \\
Thus, given a distortion function $g$, we can compute the corresponding risk measure as
\begin{align}
    \mathrm{Risk}_g(Z|\alpha) &= \int_{\mathcal{Z}} z \dd (g_{\alpha}\circ \Pr)(z)\notag \\
    &=\int_{\mathcal{Z}} g_{\alpha}(1-F_Z(z))\dd z \notag\\
    &= \int_0^1 g_{\alpha}(t)\dd q(1-t) \notag\\
    &= \int_0^1 q(1-t)\dd g_{\alpha}(t) = \int_0^1 q(1-t) g'_{\alpha}(t) \dd t.
\end{align}
Here, $q$ is the quantile function, i.e. $q(1-\alpha) = \inf\{x\geq 0|F_Z(x) \geq 1-\alpha \} = F_Z^{-1}(1-\alpha)$. This is called the Wang transform, leads us to following observations:
\begin{enumerate}
    \item For Categorical estimate of return distribution, computing risk measures will require defining a quantile function and then multiplying it with corresponding $g(\alpha_i)$ and adding over multiple $\alpha_i \in [0,1]$.
    \item For CVaR, $g_{\alpha}(t) = \min \{ \frac{t}{1-\alpha},1\}$.
    \item For EVaR, $g_{\alpha}(t) = \underset{\lambda>0}{\min}\Big\{\lambda^{-1}\ln\Big(\exp(\lambda t-\ln(1-\alpha))\Big)\Big\}$.
    \item For the Wang risk measure, $g_{\alpha}(t) = \Phi[\Phi^{-1}(t)-\Phi^{-1}(\alpha)]$.
\end{enumerate}
These observations allow us to compute the corresponding risk measures using quantile functions of the return distributions estimated using CDQNs.
Note that, every coherent risk measure can be written using the Wang transformation or distorted expectation using quantile functions if and only if there exists a monotonic concave distortion function $g_{\alpha}(t)$ corresponding to it. 

\section{Detailed Proofs}\label{sec:proof}

\begin{proof}[Proof of Theorem~\ref{thm:coherence}]
Now, let us denote the aleatory and epistemic uncertainties of a random variable $X$ as $\xi_1$ and $\xi_2$. Let us represent two coherent risk measures $U_A: \mathcal{X}|_{\xi_1} \rightarrow \mathcal{Z} \subseteq \Reals$ and $U_E: \mathcal{Z}|_{\xi_2} \rightarrow \Reals$ corresponding to distorted utility functions $U^E_{\alpha_2}$ and $U^A_{\alpha_1}$. Now, if $U_A$ and $U_E$ are coherent risk measures, we obtain 
\begin{enumerate}
    \item Monotonicity: If $U_A(X_1|_{\xi_1})=Z_1$, $U_A(X_2|_{\xi_1})=Z_2$, and $X_1|_{\xi_1} \leq X_2|_{\xi_1}$ almost surely,  $$U_E(U_A(X_1|_{\xi_1})|_{\xi_2}) = U_E(Z_1|_{\xi_2}) \leq U_E(Z_2|_{\xi_2}) = U_E(U_A(X_2|_{\xi_1})|_{\xi_2}). $$
    The inner inequality is true because if $U_A$ is coherent risk measure, then $Z_1 = U_A(X_1|_{\xi_1}) \leq U_A(X_2|_{\xi_1}) = Z_2$.
    \item Positive Homogeneity: For any $c\geq 0$,
    $$U_E(U_A(c~X|_{\xi_1})|_{\xi_2}) = U_E(c~U_A(X|_{\xi_1})|_{\xi_2}) = c~ U_E(U_A(X|_{\xi_1})|_{\xi_2}). $$
    \item Translation invariance: For any constant $a \in \Reals$,
    $$U_E(U_A(X|_{\xi_1}+a)|_{\xi_2}) = U_E(U_A(X|_{\xi_1})|_{\xi_2}+a) = U_E(U_A(X|_{\xi_1})|_{\xi_2}) + a.$$
    \item Subadditivity: For $X_1, X_2 \in \mathcal{X}$, 
    \begin{align*}
        U_E(U_A((X_1+X_2)|_{\xi_1})|_{\xi_2}) &\leq U_E((U_A(X_1|_{\xi_1})+U_A(X_2|_{\xi_1}))|_{\xi_2}) \\
        &\leq U_E(U_A(X_1|_{\xi_1})|_{\xi_2}) + U_E(U_A(X_2|_{\xi_1})|_{\xi_2}).
    \end{align*}
\end{enumerate}
Thus, composition of two coherent risk measures $U_A$ and $U_E$ quantifying the aleatory and epistemic uncertainties $\xi_1$ and $\xi_2$ is also a coherent risk measure.
\end{proof}
We observe that the existence of distorted utility functions $U^E_{\alpha_2}$ and $U^A_{\alpha_1}$ are not necessary to prove Theorem~\ref{thm:coherence}. We state the theorem statement with distorted utilities to maintain the flow of the text in the main paper.

Rather, we can leverage the observation that if a coherent risk measure $U$ also satisfies comonotonic subadditivity~\citep{SONG2009459}, we always get a concave distortion function $g_{\alpha}$ corresponding to it such that $g_{\alpha}(0)=0$ and $g_{\alpha}(1)=1$. Here,
\begin{enumerate}
    \item \textbf{Comonotonic Sub-additivity:} If $X_1$ and $X_2 \in \mathcal{X}$ are comonotonic, then $$U(X_1+X_2) \leq U(X_1) + U(X_2).$$
    \item \textbf{Comonotonicity:} Two random variables $X_1, X_2 \in \mathcal{X}$ are comonotonic, if and only if
    \begin{align*}
        [X_1(\omega_1)-X_1(\omega_2)][Y_1(\omega_1)-Y_2(\omega_2)]\geq 0
    \end{align*}
    almost surely for all $\omega_1$ and $\omega_2$ in the event space $\Omega$.
\end{enumerate}
In that case, the aforementioned proof of coherence of composite risk $U_E\circ U_A$ naturally extends for the composition of corresponding distorted utility functions $U^E_{\alpha_2}$ and $U^A_{\alpha_1}$.

\begin{remark}\label{remark:dual}
	If the random variable $Z$ follows a distribution $P$, any coherent risk measure $U: \Delta(\mathcal{Z}) \rightarrow \Reals$ can be written in a dual form: $\mathrm{Risk}_{1-\alpha}(Z|P) = \sup_{Q \in Q_\alpha} \mathbb{E}[Z]$. Here, $Q_\alpha$ is a set of distributions defined around $P$ constrained by $\alpha$ and on support of $P$. For example, in case of CVaR, $Q_\alpha = \{Q \ll P: \frac{\dd Q}{\dd P} \leq \frac{1}{\alpha} \text{ almost surely}\}$, and in case of Entropic VaR~\citep{ahmadi2012entropic}, $Q_\alpha = \{Q\ll P: D_{KL}(Q||P) \leq -\ln \alpha\}$. For $\alpha =1$, the risk measures reduce to expectation and $Q_{\alpha=1}= \{P\}$.
\end{remark}

\begin{proof}[Proof of Theorem~\ref{thm:comp_geq_add}]
	Let $F^C(U_A, U_E, \beta) \defn U_E(U_A(X|_{\xi_A})|_{\xi_E})$ be the primal form of composite risk, and the dual form is: $F^C(U_A, U_E, \beta) \triangleq \sup_{\beta' \in \Beta_{\alpha_2}} \mathbb{E}_{\theta \sim \beta'}\left[\sup_{Q\in Q^{\theta}_{\alpha_1}} \mathbb{E}_{Z\sim Q(.|\theta)}[Z]\right].$ 
	
	\textit{Part a:} By replacing the variance with a dual of a coherent risk measure in~\citep{clements2019estimating}, we obtain: 
	\begin{align*}
	    F^A(U^A, \beta) &= A(U^A, \beta) + \sup_{Q\in Q^{\hat{\theta}}_{\alpha}}\int_{\Theta}\int_{\mathcal{Z}}z \dd Q(z|\theta)\dd\beta(\theta)\\ &=\mathbb{E}_{\theta \sim \beta}\left[\sup_{Q\in Q^{\theta}_{\alpha}} \mathbb{E}_{Z\sim Q(.|\theta)}[Z]\right]\\
	    &= F^C(U_A, I, \beta).
	\end{align*}
	The penultimate equality is obtained by the centered definition of aleatory risk. The last inequality is a direct consequence of the definition of the composite risk.
	
	\textit{Part 2:} The second claim follows from Remark~\ref{remark:dual}. First, we observe that $\beta \in \Beta_\alpha$ as $\Beta_\alpha = \{\beta' \ll \beta \, | \, f_1\Big(\frac{\dd \beta'}{\dd \beta}\Big) \leq f_2(\alpha)\}$ and $f_1\Big(\frac{\dd P}{\dd P}\Big) = 0 \leq f_2(\alpha)$ for any $\alpha \in [0,1]$.
	Now, let us denote
 	\begin{align*}
		\beta^{*} &\triangleq \underset{\beta' \in \Beta_\alpha}{\arg\sup} \, \mathbb{E}_{\beta'} [-Z |_{\xi_E}].
	\end{align*}
	Thus, if $\beta^{*} \neq \beta$, then $F^C(U_A, U_E, \beta) \geq F^C(U_A, I, \beta) = F^A(U_A, \beta)$.
	We conclude the proof by observing that for $\alpha \neq 1$, $\beta^{*} \neq \beta$.
 
\end{proof}
The aforementioned proof is independent of the existence of distortion function. If it exists for the epistemic risk measure, the proof is even straightforward.
If we assume that there exists a distortion function $U^E_{\alpha_2}$ for epistemic risk, we get $U^E_{\alpha_2}(t) \geq t$ for all $t\in [0,1]$. Because $U^E_{\alpha_2}$ is \textit{concave}, and $U^E_{\alpha_2}(0)=0$ and $U^E_{\alpha_2}(1)=1$. Thus, it will be almost always above $t$ by definition of concave function.
\clearpage

\section{Additional Experimental Results}

In this section we provide additional experiment results that did not make it into the main paper. These involve testing the distributional fit of the return distribution using two different distributional RL frameworks, an empirical experiment demonstrating why composite risk is preferable over additive risk and 
\subsection{Effect of Ensemble Size on Performance and Computation Time}\label{sec:ensemble_size}

In the following experiment we investigate how the number of estimators in the ensemble affect the results and running time of the algorithm.

\begin{table*}[h!]
	\centering
		\caption{Performance of risk-neutral SENTINEL-K in the \emph{CartPole-v0} environment. Shown is number of \emph{falls} (lower is better) and the time elapsed per experiment in seconds. The results were taken over $20$ independent runs for each ensemble size. $\sigma$ is the standard error of the mean.}\label{tab:ensemble_size}
		\begin{tabular}{c|c|c}
			Ensemble size & $\#$ Falls $\pm\sigma$& Time elapsed (s) per experiment $\pm\sigma$\\
			\hline
			$K=1$& $5332.4\pm404.86$ & $469.8\pm14.2$\\
			$K=2$& $4627.8\pm386.2$ & $1886.3\pm130.6$\\
			$K=4$& $4357.9\pm334.4$ & $4285.6\pm204.5$\\
			$K=8$& $3532.8\pm207.5$ & $16528.2\pm1479.9$
	\end{tabular}
\end{table*}

In Table~\ref{tab:ensemble_size} we can see a monotonic increase in performance with the size of the ensemble. However, with each added estimator we can also observe a sizeable increase in computation time per experiment. Thus, there is a trade-off between computation time and performance and while more estimators would be preferable to use we chose to use $K=4$ for most of the experiments. 

\subsection{Return Distribution Estimation}

In these experiments, we verify the goodness of fit of the used DRL framework (CDQN) and compare the results with another DRL framework (VDQN).

\begin{figure*}[h!]
	\centering
	\begin{subfigure}[b]{0.24\textwidth}
		\centering
		\includegraphics[width=\textwidth]{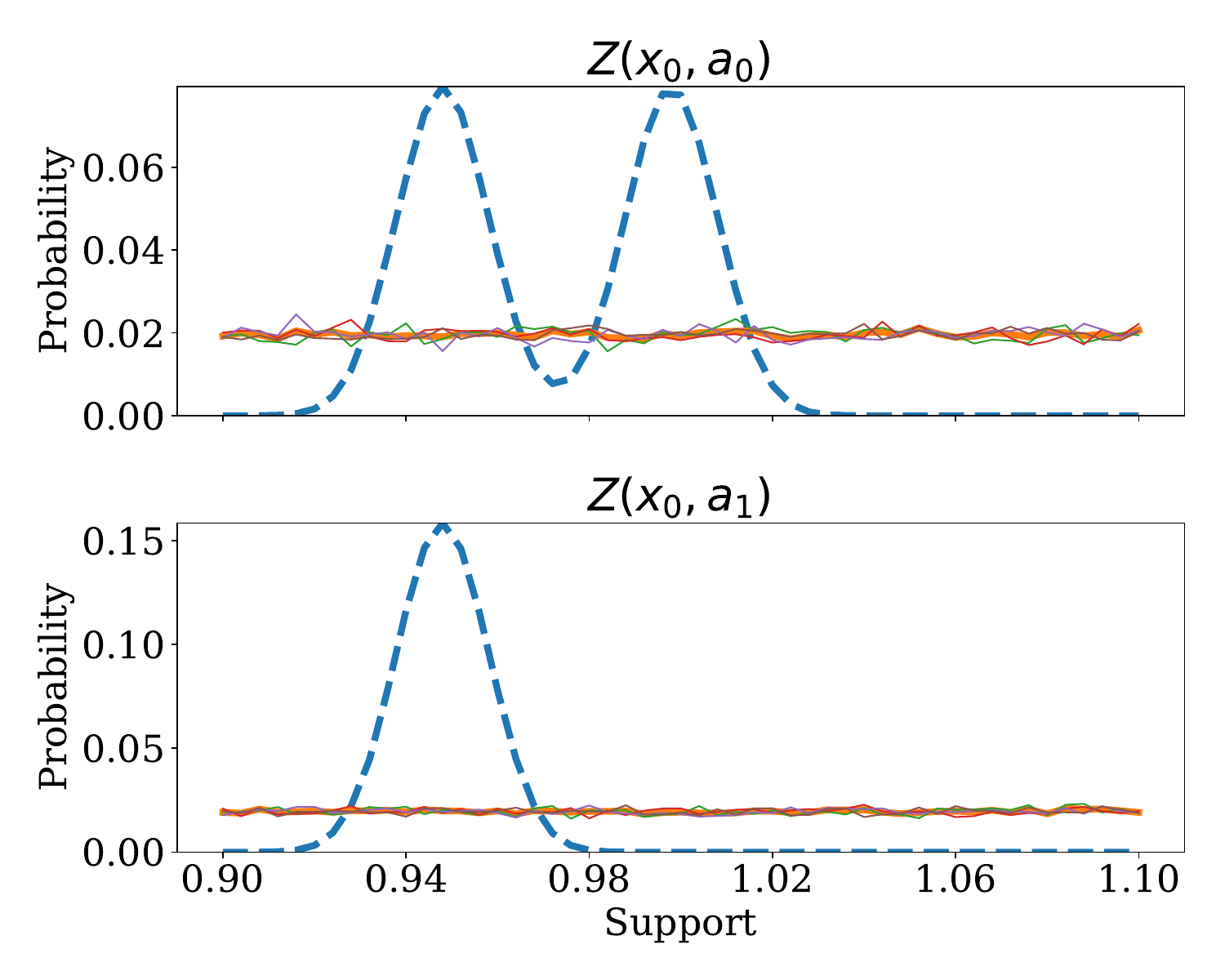}
		\caption{$n=0$}
	\end{subfigure}
	\hfill
	\begin{subfigure}[b]{0.24\textwidth}
		\centering
		\includegraphics[width=\textwidth]{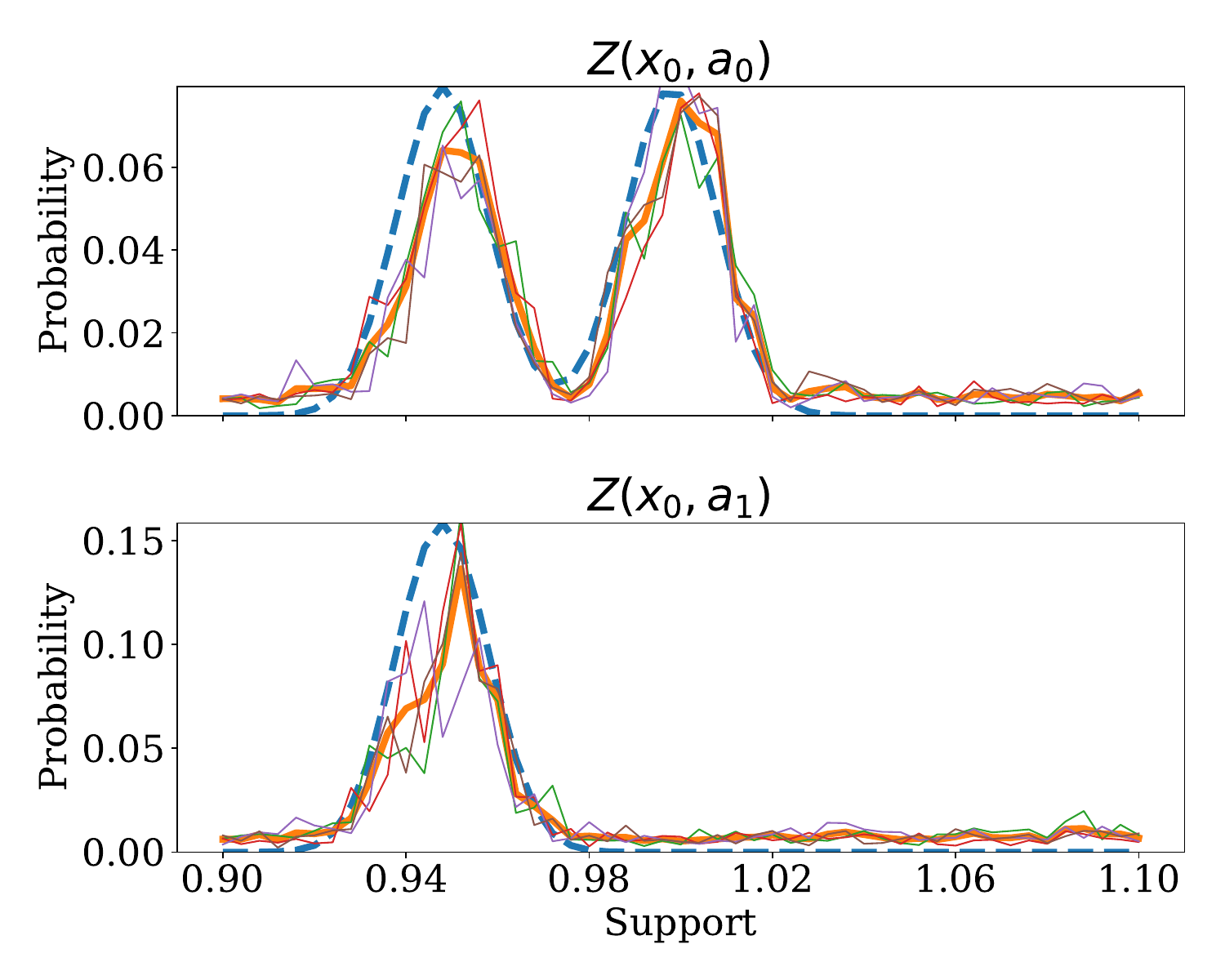}
		\caption{$n=1000$}
	\end{subfigure}
	\hfill
	\begin{subfigure}[b]{0.24\textwidth}
		\centering
		\includegraphics[width=\textwidth]{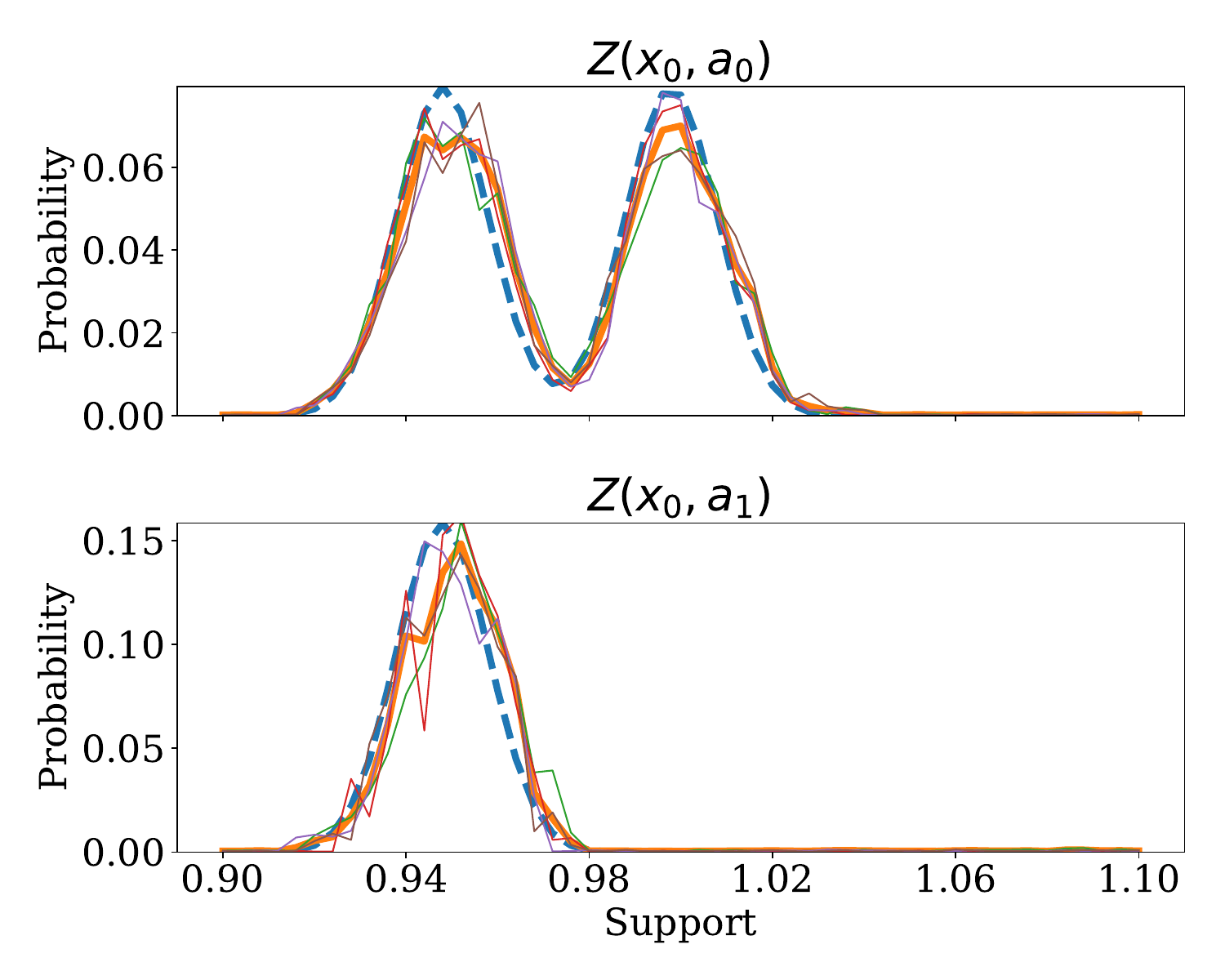}
		\caption{$n=5000$}
	\end{subfigure}
	\hfill
	\begin{subfigure}[b]{0.24\textwidth}
		\centering
		\includegraphics[width=\textwidth]{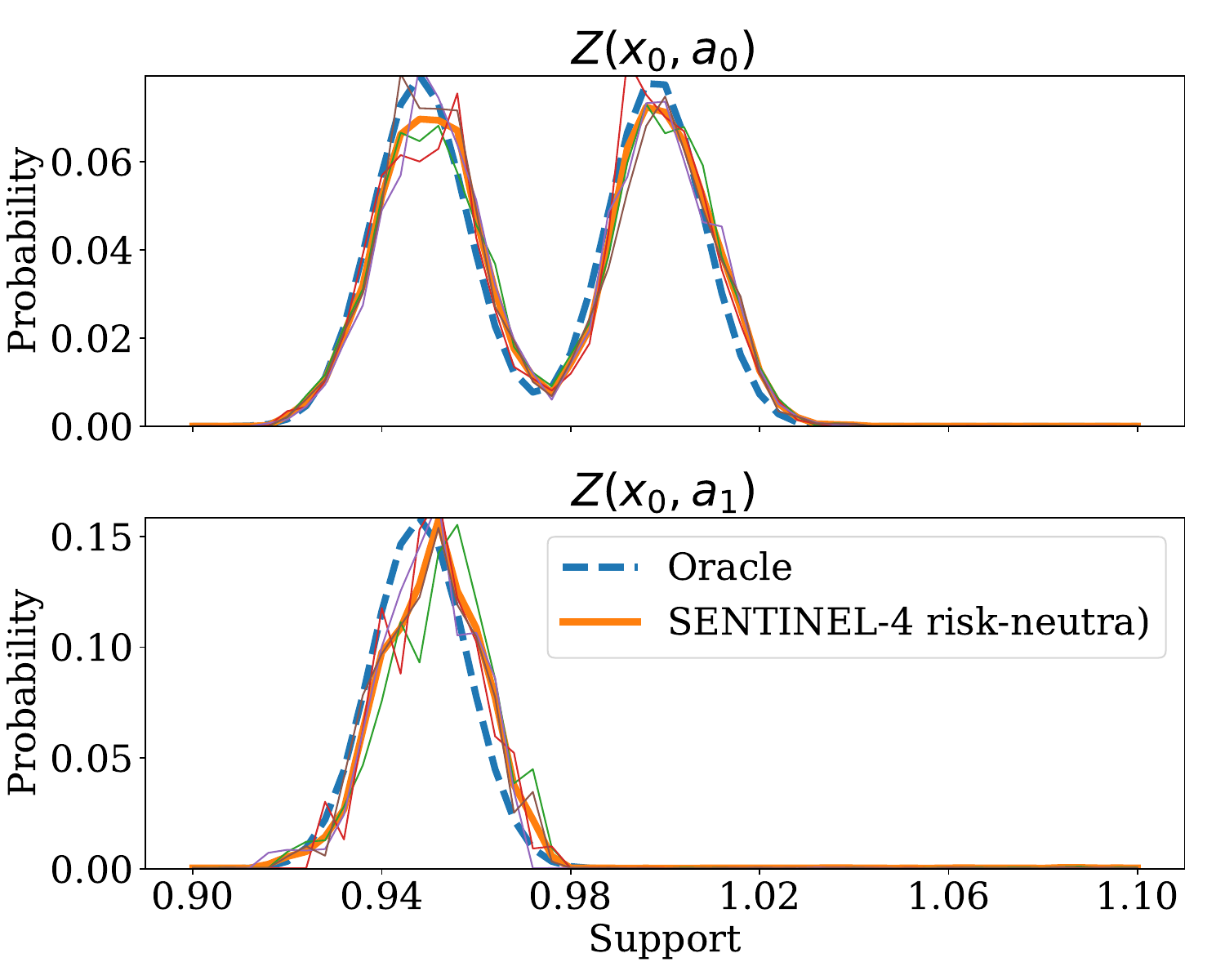}
		\caption{$n=10000$}
	\end{subfigure}
	\caption{Return distributions of $a_0$ and $a_1$ for $0, 1000, 5000$ and $10000$ data points ($n$) respectively. The blue dashed line is the categorical approximation of $Z(s_0, a_0)$ and $Z(s_0, a_1)$ respectively. The thick orange line is the marginal posterior $\Pr(\hat{Z})$ with SENTINEL-4. The thin lines are the posteriors of the individual estimators.}	\label{fig:toy_example1}
\end{figure*}
\begin{figure}[h!]
\centering
	\includegraphics[width=0.5\textwidth]{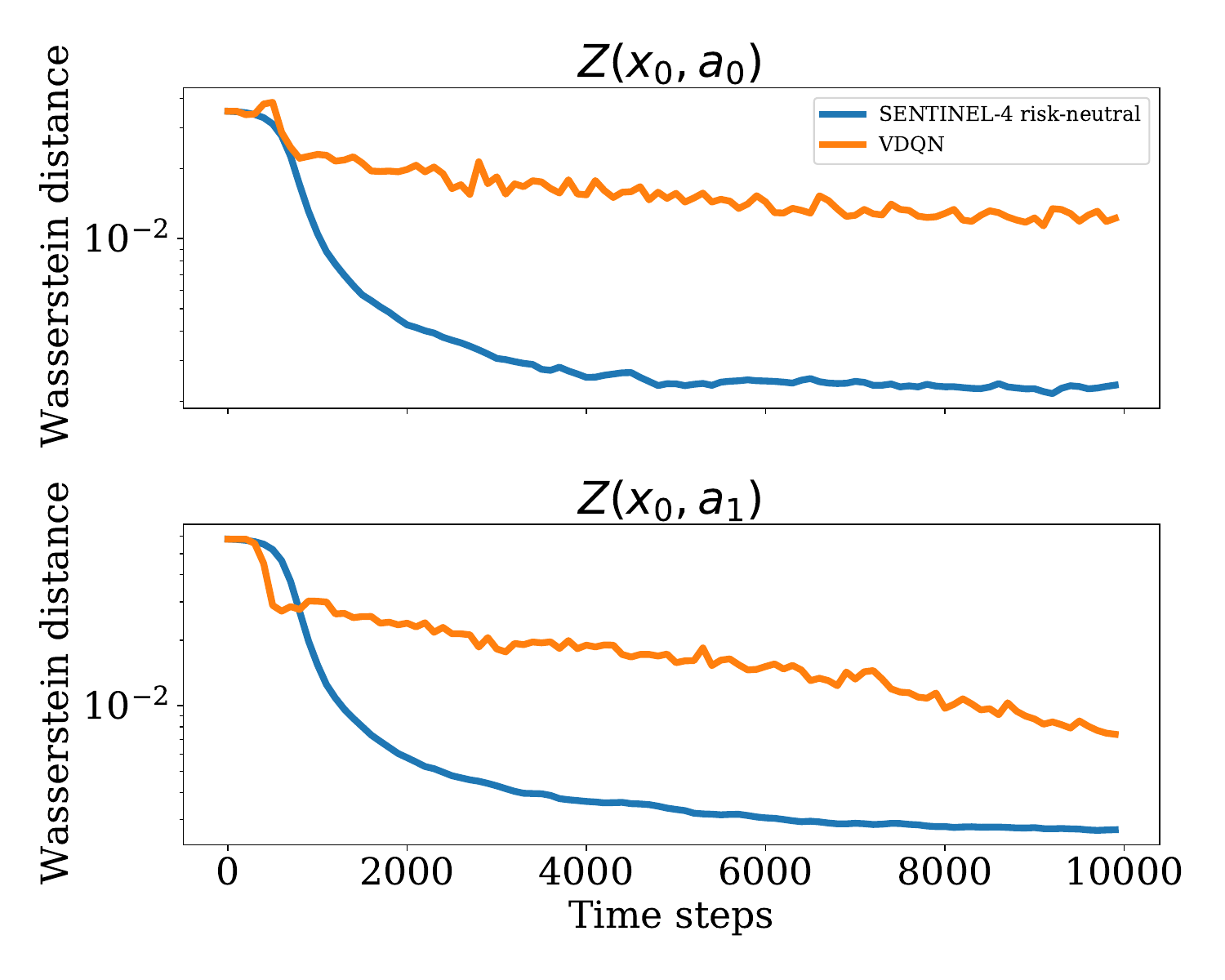}%
	\caption{Shows convergence in distribution of SENTINEL-4 (risk-neutral) and VDQN by measuring the Wasserstein distance between the categorical approximation of $Z(s_0, a_0), Z(s_0, a_1)$ and the estimated distributions by the two agents, for each action.}
	\label{fig:toy_example2}
\end{figure}

In order to demonstrate uncertainty estimation and convergence in distribution of SENTINEL-K framework, we test SENTINEL-4 on an MDP environment with known multimodal return distribution.
The MDP contains three states and two actions such that the return distribution of $a_0$ from state $s_0$ is a mixture of Gaussians $Z(s_0, a_0) \sim \sum_{i=0}^N \Phi_i \mathcal{N}(\mu_i, \sigma_i)$ and the return distribution of action $a_1$ is  $Z(s_0, a_1) \sim \mathcal{N}(\mu_1, \sigma_1)$. Here, $\Phi = [0.5, 0.5]$, $\mu = [1.0, 0.95], \sigma = [0.1, 0.1]$. 
Figure~\ref{fig:toy_example1} shows convergence in distribution of SENTINEL-4. We observe that SENTINEL-4 estimates the return distributions of both the actions considerably well after using $5000$ data points.

In Figure~\ref{fig:toy_example2}, we further illustrate the Wasserstein distance of the distributions estimated by risk-neutral SENTINEL-4 and VDQN algorithms from the true return distribution. We show that the VDQN fails to converge to the true return distribution whereas SENTINEL-4 converges to the true return distribution in significantly less number of steps.

\subsection{Composite Risk vs. Additive Risk}\label{sec:app_compvsadditive}

\begin{figure}[h!]
	\centering
	\includegraphics[width=0.5\textwidth]{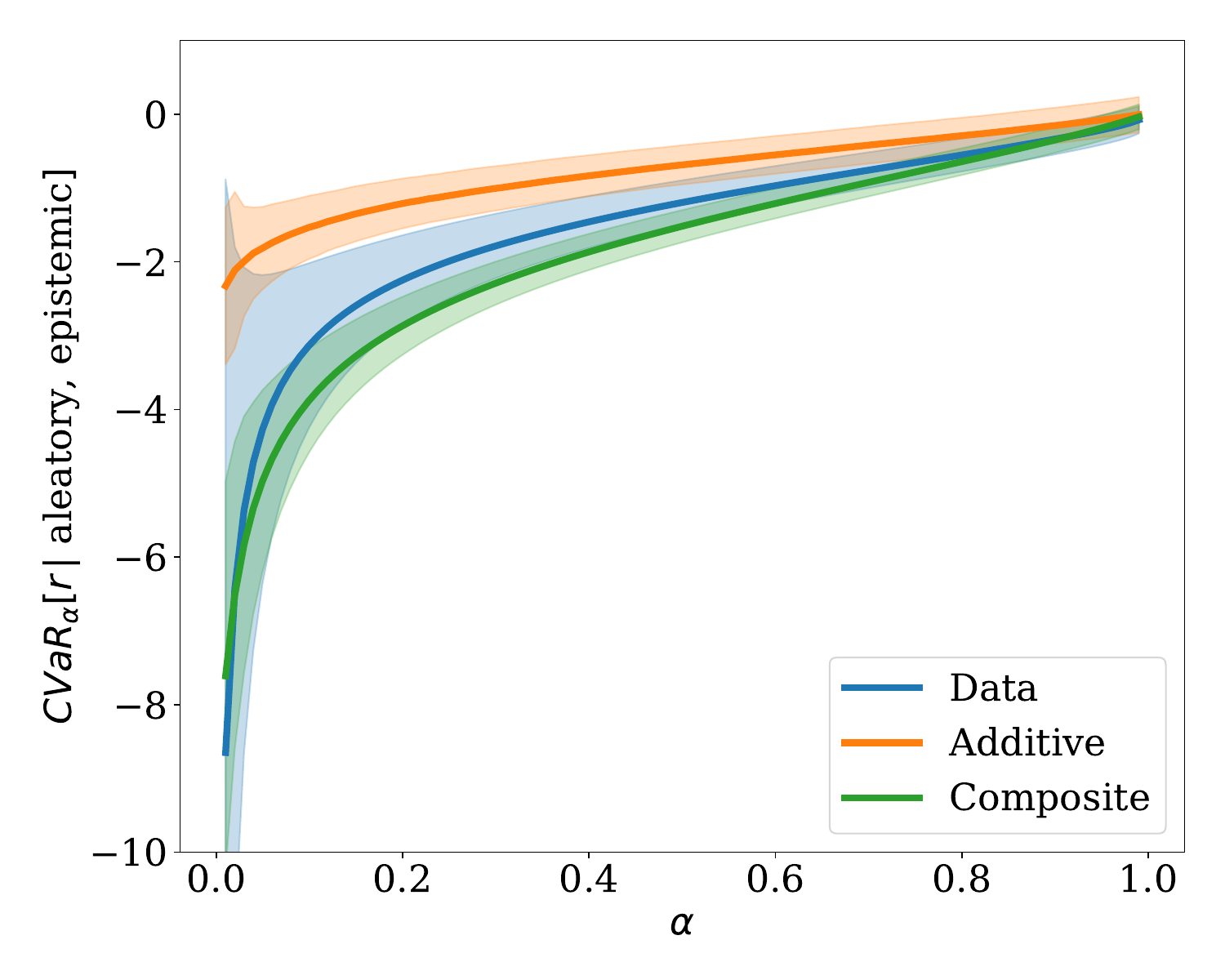}\\%
	\caption{Estimation of total $CVaR_{\alpha}$ from a mixture of 100 Gaussians sampled from a posterior distribution. Total $CVaR_{\alpha}[Data]$ is based on the marginal distribution of the $r$ as given in Example~\ref{ex:gauss-mix-cvar2}. We compare this with composite and additive estimates and illustrate results over $100$ runs. Here, lower value of CVaR indicates higher mass on the left tail of the distribution and thus, higher risk of obtaining low returns.}
    \label{fig:gauss-mix-cvar2}
\end{figure}

In order to compare the risk estimation using additive and composite formulations, we consider an example of estimating CVaR over a Gaussian mixture.
\begin{example}
  \label{ex:gauss-mix-cvar2}
    We consider a mixture of $100$ Gaussians: $p(r) = \sum_{i=1}^{100} \phi_i\mathcal{N}(\mu_i, \sigma_i^2)$, where $\Phi \sim Dir([0.5]^{100}), \mu \sim \mathcal{N}(0, 1)$, and $\sigma^2 \sim \Gamma^{-1}(2, 0, 1)$.
	We compute $CVaR_{\alpha}[r]$ from the data generated from such mixture for 100 runs. We further estimate composite risk with $U_E, U_A = CVaR_{\alpha}$ and additive risk with $U_A= CVaR_{\alpha}$. The results illustrated in Figure~\ref{fig:gauss-mix-cvar2} show that the additive CVaR risk strictly underestimates the total CVaR risk computed from the data, whereas the composite risk is closer to the one computed from data. Specifically, for lower values of $\alpha$ (specifically, $\alpha \leq 0.5$), i.e. towards the extreme end of the left tail where events occur with low probability, the additive CVaR risk deviates significantly from data whereas the composite measure yields closer estimation. Such values of $\alpha$'s are typically interesting for risk-sensitive applications.
\end{example}

In the following example the sensitivity w.r.t the parameter uncertainty of the composite risk formulation is shown. As the belief concentrates, both the composite and additive risk formulations ends up with the optimal behaviour for this problem, as seen in Figure~\ref{fig:gauss-mix-cvar2}. The main difference in behaviour arises in situations with high parameter uncertainty.

\begin{example}[Composite vs. Additive risk for Gaussian estimators]
Let $Z(z;s_1, a_1) = \mathcal{N}(z;\theta_1, \theta_2), Z(z;s_1, a_2) = \omega\mathcal{N}(z;\theta_3, \theta_4)+(1-\omega)\mathcal{N}(z;\theta_5, \theta_6)$ for $\omega \in [0, 1]$. Let $\omega \sim Beta(\theta_7, \theta_8)$. Then, let $\theta=[0, 1, 1, 1, -1, 1, 1, 1]^\top$. In Figure~\ref{fig:gauss-mix-cvar2} the example is illustrated. 
\end{example}

\begin{figure}[h!]
\centering
	\includegraphics[width=0.5\textwidth]{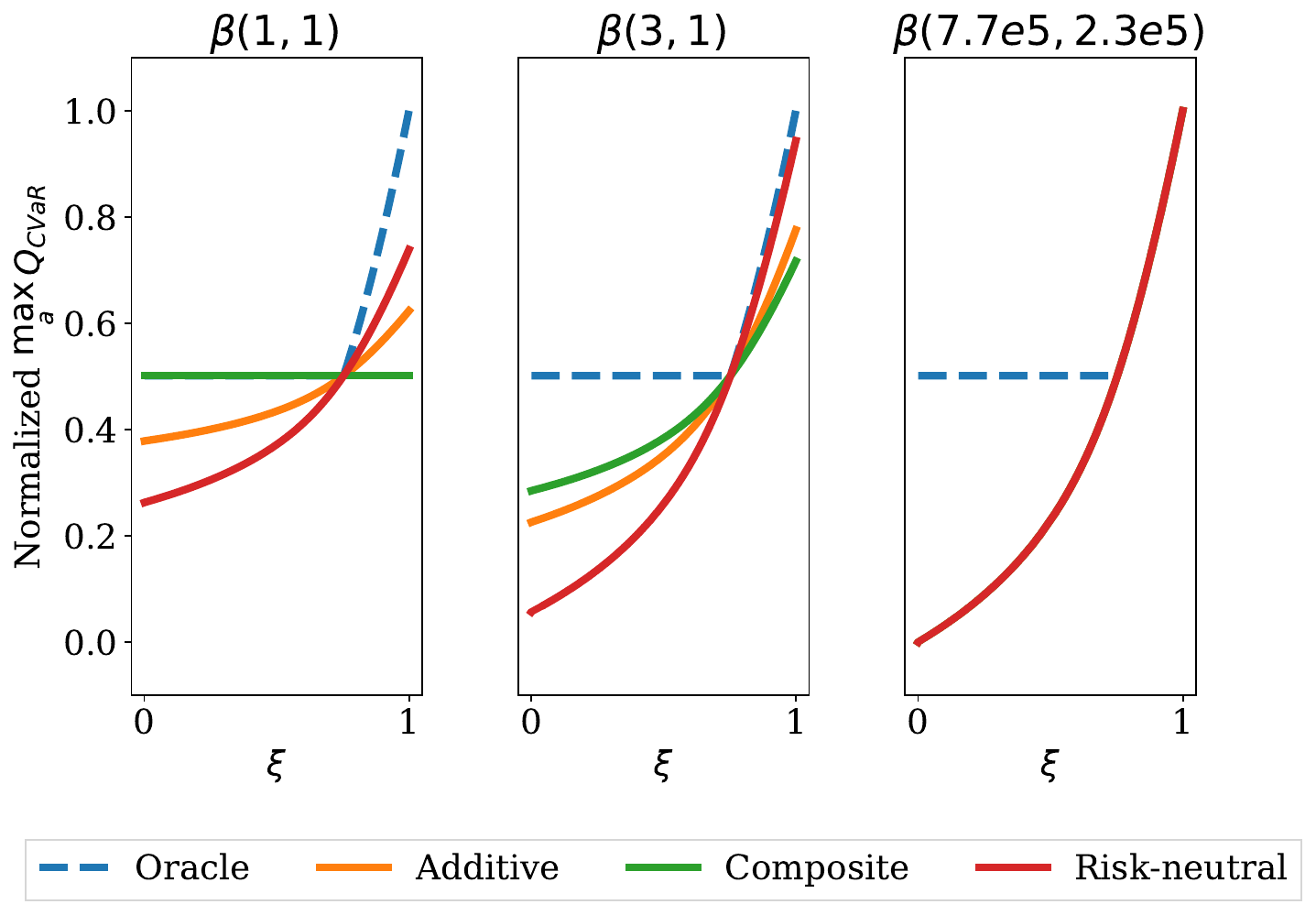}%
	\caption{Illustrates the behaviour of a risk-neutral decision-maker and two risk-sensitive decision-makers, with additive and composite risk respectively. Shown in the figure is the normalized $\underset{a}{\max} \, Q_{CVaR}$ for $\alpha=0.25$.}
	\label{fig:distributions}
\end{figure}

\section{Additional Details}

In this section we further describe the experimental details and the parameters chosen for said experiment.

\subsection{Data Masking}\label{sec:datamask}

Similar to~\citep{osband2016deep}, we use data masking to ensure the estimators have access to different parts of the data. The authors in that paper mention a few ways of doing this, namely using a Bernoulli mask, an exponential mask and a Poisson mask. In this work we chose to use a Bernoulli mask with parameter $p=\frac{1}{3}$. This means that in expectation, each estimator has access to a third of the full data set $\mathcal{D}$.

Upon observing a transition $s, a, r, s'$, we sample $K$ parameters from $Ber(\frac{1}{3})$, and assign those parameters to each estimator, respectively, for that particular transition. As an example, consider the following table in Table~\ref{tab:datamask}, where the columns of the data $\mathcal{D}$ has been augmented with the Bernoulli mask.

In this example, the first estimator will have the first and the last transition available to it, since $m_{1, t} = m_{1, T} = 1$, while the second and the $K$'th estimator will not have access to the first transition, since $m_{2, t} = m_{K, t} = 0$.

\begin{table*}[h!]
	\centering
		\caption{An example of data masking on transitions $[\tau_t, \tau_T]$, where $s_t$ denotes the state at time $t$, $a_t$ the action taken at time $t$, $r_t$ the reward received at time $t$ and $s_{t+1}$ the successor state of $s_t$ at time $t$. $m_{1, t}$ denotes the availability of transition $t$ to the first estimator.}\label{tab:datamask}
		\begin{tabular}{c|c|c|c|c|c|c|c}
			$s$ & $a$ & $r$ & $s'$ & $m_1$ & $m_2$ & $\cdots$ & $m_K$\\
			\hline
			$s_t$& $a_t$ & $r_t$ & $s_{t+1}$ & $1$ & $0$ & $\cdots$ & $0$\\
			$s_{t+1}$& $a_{t+1}$ & $r_{t+1}$ & $s_{t+2}$ & $0$ & $0$ & $\cdots$ & $1$\\
			$\cdots$& $\cdots$ & $\cdots$ & $\cdots$ & $\cdots$ & $\cdots$ & $\cdots$ & $\cdots$\\
			$s_{T-1}$& $a_{T-1}$ & $r_{T-1}$ & $s_{T}$ & $1$ & $1$ & $\cdots$ & $1$\\
	\end{tabular}
\end{table*}

\subsection{Addendum on \emph{Follow the Regularised Leader}}\label{sec:appendix_ftrl}

Follow the regularised leader in our setting can be seen as a mapping $f_\lambda : \mathcal{X}^K \rightarrow \mathcal{X}$, where the mapping is taken over a convex combination over the $K$ densities. Let $f_k(x)$ be the $k$'th probability density function over $x$, then,
\begin{align*}
    f_\lambda(x) &= \sum_{k=1}^K w_k f_k(x)\\
    \text{where, } ||\mathbf{w}||_1 &= 1 \wedge \forall_k w_k \geq 0.
\end{align*}

\definecolor{tableau101}{RGB}{31, 119, 180}
\definecolor{tableau102}{RGB}{255, 127, 14}
\definecolor{tableau103}{RGB}{44, 160, 44}
\definecolor{tableau104}{RGB}{214, 39, 40}
\definecolor{tableau105}{RGB}{148, 103, 189}
\definecolor{tableau106}{RGB}{140, 86, 75}
\definecolor{tableau107}{RGB}{227, 119, 194}

\begin{figure}[h!]
    \centering
    \def\mmuone{0}
    \def\mmutwo{-2}
    \def\mmuthr{2}
    \def\sstdone{1}
    \def\sstdtwo{0.5}
    \def\sstdthr{1}

    \begin{subfigure}{0.48\textwidth}
        \begin{tikzpicture}[scale=1, x=0.5cm, y=4cm]
        \draw[->, line width=1pt] (-6, 0) -- (6.0, 0) node[right] {$R$};
        \draw[->, line width=1pt] (0, 0) -- (0.0, 0.6) node[above] {$p$};
        \draw[scale=1.0, domain=-6:6, smooth, variable=\x, tableau101, line width=2pt] plot ({\x}, {1/(\sstdone*sqrt(2*pi)) * e^(-1/2 * ((\x-\mmuone)/\sstdone)^2)}) {};
        \draw[scale=1.0, domain=-6:6, smooth, variable=\x, tableau102, line width=2pt] plot ({\x}, {1/(\sstdtwo*sqrt(2*pi)) * e^(-1/2 * ((\x-\mmutwo)/\sstdtwo)^2)}) {};
        \draw[scale=1.0, domain=-6:6, smooth, variable=\x, tableau103, line width=2pt] plot ({\x}, {1/(\sstdthr*sqrt(2*pi)) * e^(-1/2 * ((\x-\mmuthr)/\sstdthr)^2)}) {};
      \end{tikzpicture}
      \caption{Estimator posteriors $f_k(x)$}
    \end{subfigure}
    \begin{subfigure}{0.48\textwidth}
        \begin{tikzpicture}[scale=1, x=0.5cm, y=4cm]
        \draw[->, line width=1pt] (-6, 0) -- (6.0, 0) node[right] {$R$};
        \draw[->, line width=1pt] (0, 0) -- (0.0, 0.6) node[above] {$p$};
        \draw[scale=1.0, domain=-6:6, smooth, variable=\x, tableau104, line width=2pt, dashed] plot ({\x}, {
          1/3 * 1/(\sstdone*sqrt(2*pi)) * e^(-1/2 * ((\x-\mmuone)/\sstdone)^2) + 
          1/3 * 1/(\sstdtwo*sqrt(2*pi)) * e^(-1/2 * ((\x-\mmutwo)/\sstdtwo)^2) + 
          1/3 * 1/(\sstdthr*sqrt(2*pi)) * e^(-1/2 * ((\x-\mmuthr)/\sstdthr)^2)}) {};
      \end{tikzpicture}
      \caption{Marginal over estimators $\frac{1}{K}\sum_k f_k(x)$}
    \end{subfigure}
    \begin{subfigure}{0.48\textwidth}
        \begin{tikzpicture}[scale=1, x=0.5cm, y=4cm]
        \draw[->, line width=1pt] (-6, 0) -- (6.0, 0) node[right] {$R$};
        \draw[->, line width=1pt] (0, 0) -- (0.0, 0.6) node[above] {$p$};
        \draw[scale=1.0, domain=-6:6, smooth, variable=\x, tableau105, line width=2pt] plot ({\x}, {
        1/3 * 1/(\sstdone*sqrt(2*pi)) * e^(-1/2 * ((\x-\mmuone)/\sstdone)^2) + 
        1/3 * 1/(\sstdtwo*sqrt(2*pi)) * e^(-1/2 * ((\x-\mmutwo)/\sstdtwo)^2) + 
        1/3 * 1/(\sstdthr*sqrt(2*pi)) * e^(-1/2 * ((\x-\mmuthr)/\sstdthr)^2)}) {};
    \end{tikzpicture}
    \caption{FTRL($\lambda=0$)}
  \end{subfigure}
  \begin{subfigure}{0.48\textwidth}
    \begin{tikzpicture}[scale=1, x=0.5cm, y=4cm]
      \draw[->, line width=1pt] (-6, 0) -- (6.0, 0) node[right] {$R$};
      \draw[->, line width=1pt] (0, 0) -- (0.0, 0.6) node[above] {$p$};
      \draw[scale=1.0, domain=-6:6, smooth, variable=\x, tableau106, line width=2pt] plot ({\x}, {1/(\sstdtwo*sqrt(2*pi)) * e^(-1/2 * ((\x-\mmutwo)/\sstdtwo)^2)}) {};
    \end{tikzpicture}
    \caption{FTRL($\lambda\rightarrow\infty$)}
  \end{subfigure}
  \begin{subfigure}{0.48\textwidth}
    \begin{tikzpicture}[scale=1, x=0.5cm, y=4cm]
        \draw[->, line width=1pt] (-6, 0) -- (6.0, 0) node[right] {$R$};
        \draw[->, line width=1pt] (0, 0) -- (0.0, 0.6) node[above] {$p$};
        \draw[scale=1.0, domain=-6:6, smooth, variable=\x, tableau107, line width=2pt] plot ({\x}, {
        0.26 * 1/(\sstdone*sqrt(2*pi)) * e^(-1/2 * ((\x-\mmuone)/\sstdone)^2) + 
        0.4 * 1/(\sstdtwo*sqrt(2*pi)) * e^(-1/2 * ((\x-\mmutwo)/\sstdtwo)^2) + 
        0.34 * 1/(\sstdthr*sqrt(2*pi)) * e^(-1/2 * ((\x-\mmuthr)/\sstdthr)^2)}) {};
    \end{tikzpicture}
    \caption{FTRL($\lambda=1$)}
  \end{subfigure}
    \caption{Shows how FTRL transforms a set of probability distributions into a weighted average, with the weights depending on the regularising parameter $\lambda$. In (a), three Gaussian distributions are given and in (b), the marginal distribution over returns can be seen, having marginalised out the estimators. In (c, d, e) the weighted distributions can be seen when varying $\lambda$. If $\lambda=0$, as in (c), then the weighted distribution is the same as in (b). If $\lambda\rightarrow \infty$ as in (d), then the estimator that is closest to the marginal (using Kullback-Leibler divergence) will be assigned all weight. Finally, for $0 \leq \lambda < \infty$ the weighting is somewhere in between \emph{Average} and \emph{Greedy} selection.}
    \label{fig:ftrl}
\end{figure}
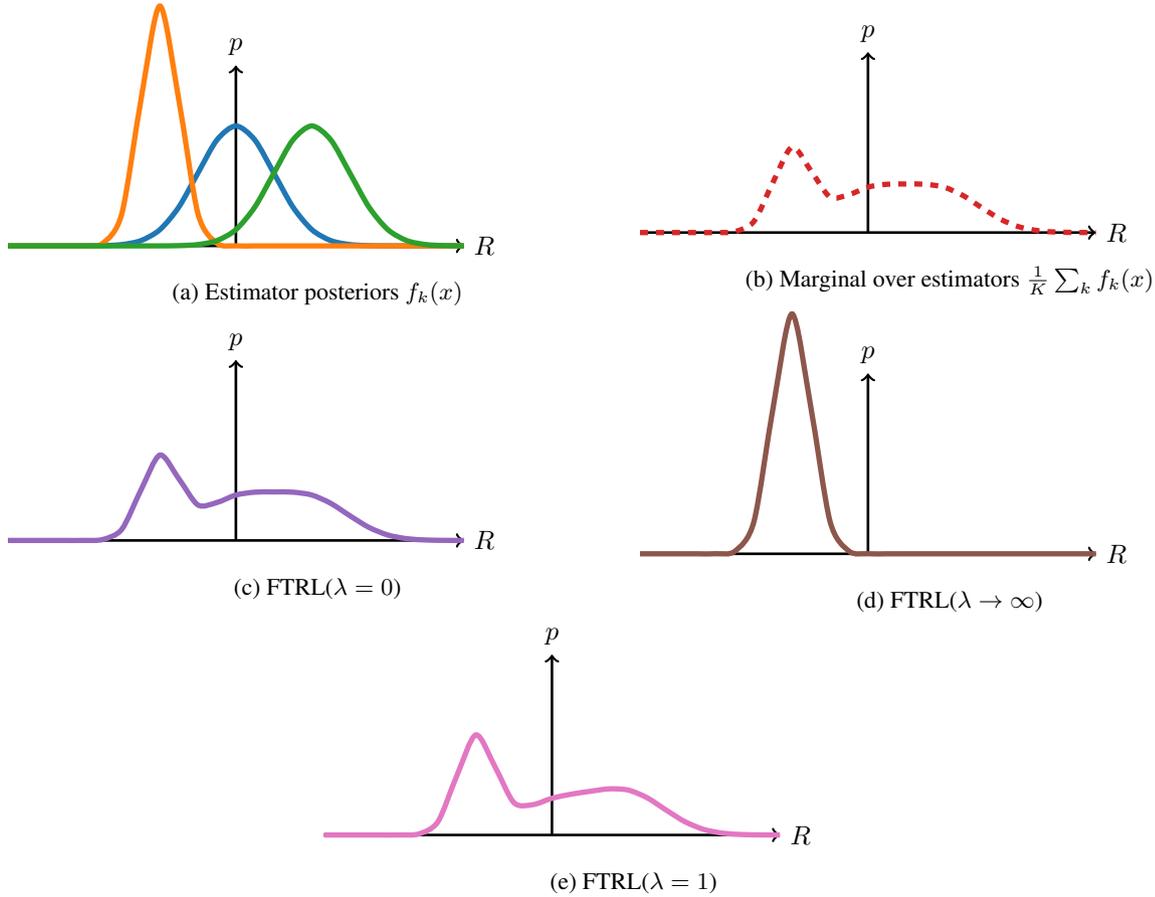

Consider the following example with Gaussian estimators.
\begin{example}[FTRL with Gaussian estimators]
    Let $f_k(x) = \mathcal{N}(\mu_k, \sigma^2_k), \mu = [0, -2, 2], \sigma = [1, 0.5, 1]$, respectively. We can compute the Kullback-Leibler divergence from each of the estimators to the marginal experimentally, and get approximately the following, $D_{KL}\Big(\frac{1}{K}\sum_k f_k(x) \, || \, f_i(x)\Big) = [0.48, 0.92, 0.74]$. Now, using the exponentiated FTRL approach as defined in Section~\ref{sec:algo}, we get that $w = \exp\Big([0.48\lambda, 0.92\lambda, 0.74\lambda]\Big)/(w_1+w_2+w_3)$. This leads to $w = [0.26, 0.40, 0.34]$ for $\lambda=1$ as used in most of the experiments. If $\lambda= 0$, then $w = [\frac{1}{3}, \frac{1}{3}, \frac{1}{3}]$. Finally, if $\lambda \rightarrow \infty$, then $w = [0, 1, 0]$, which assigns all probability to the estimator that is the most similar to the marginal distribution. The distributions given from FTRL, the original estimators and the marginal distribution can be seen in Figure~\ref{fig:ftrl}.
\end{example}

\subsection{Compute Specifications and Total Compute}\label{sec:compute}
In this section we explain the specifications of the computers the experiments were ran in, and compare the compute time for the different algorithms. In Table~\ref{tab:alg_comp} the compute time is shown for CDQN~\citep{bellemare2017distributional}, UA-DQN~\citep{clements2019estimating}, VDQN~\citep{tang2018exploration} and the proposed algorithm in this paper.

The majority of the computations were ran on NVIDIA Tesla T4 GPUs with 16 GB RAM and 16 core Intel(R) Xeon(R) Gold 6226R CPUs @ 2.90GHz with 768GB DDR4 RAM and the hyperparameters were selected such that the shared hyperparameters are the same and with the algorithm specific parameters chosen such that the overall compute time is similar in most cases. 
\begin{table*}[ht!]
	\centering
		\caption{Compute time for the different algorithms in the \emph{Highway-v1} environment. Shown is the time elapsed per experiment in seconds.}\label{tab:alg_comp}
		\begin{tabular}{l|c}
			Algorithm & Time elapsed (s) per experiment\\
			\hline
			CDQN& $\approx 10614$\\
			UA-DQN& $\approx 13748$\\
			VDQN& $\approx 13069$\\
			SENTINEL-4& $\approx36384$\\
	\end{tabular}
\end{table*}

\newpage
\section{Hyperparameters}\label{sec:hyperparam}

In this section we show the hyperparameters used for the different experiments, including problem parameters, algorithm parameters and network structure. The choice of hyperparameters is firstly done to match the shared parameters, (such as minibatch size, update schedules and ensemble size), then secondly to match the number of learnable parameters for the model. Finally, we attempt to match the computation time.

\subsection{FTRL vs. Average. vs Greedy.}

The experimental results for this experiment can be seen in Figure~\ref{fig:cartpole}, where the \emph{follow the regularised leader} parameter $\lambda$ is varied across experiments.

\begin{table*}[h!]
	\centering
		\caption{Hyperparameters for the FTRL vs. Average vs. Greedy experiment.}\label{tab:hyperparam1}
		\begin{tabular}{l l r}
		\hline
		& Hyperparameter & Value\\
		\hline
		\emph{Problem parameters} & &\\
		& Environment & \emph{CartPole-v0}\\
		& State dimensions $|\mathcal{S}|$ & $4$\\
		& Action dimensions $|\mathcal{A}|$ & $2$\\
		& Maximum episode length & $200$\\
		\hline
		\emph{Algorithm parameters} & &\\
		& Discount $\gamma$ & $0.99$\\
		& Number of atoms & $51$\\
		& Maximum steps in env & $1e5$\\
		& Initial $\epsilon$ & $1.0$\\
		& Final $\epsilon$ & $0.05$\\
		& Samples from replay buffer & $100$\\
		& Replay buffer size & $1e5$\\
		& Minibatch size & $32$\\
		& Regularising parameter $\lambda$ & $[0.01, 0.1, 1.0, 4.6]$\\
		& Return distribution range $[V_{min}, V_{max}]$ & $[0, \frac{1-\gamma^{200}}{1-\gamma}]$\\
		& Update ensembles at steps & $[100, 200, \cdots]$\\
		& Update target ensembles at steps & $[1000, 2000, \cdots]$\\
		& Ensemble size $K$ & $4$\\
		& Learnable parameters & $(32 \, |\mathcal{S}| + 11891)K |\mathcal{A}|$\\
		& Optimiser & \emph{Adam}\\
		& Learning rate & $0.00025$\\
		& Network structure & $\underset{\text{input}}{4} \rightarrow \underset{\text{dense}}{32} \rightarrow \underset{\text{dense}}{32} \rightarrow \underset{\text{dense}}{128} \rightarrow \underset{\text{output}}{(51, 2)}$\\
		\hline
	\end{tabular}
\end{table*}

\clearpage
\subsection{Effect of Ensemble Size on Performance and Computation Time}

In Table~\ref{tab:ensemble_size} the results are shown for varying the ensemble size. In this section, we hyperparameters used for that experiment are displayed.

\begin{table*}[h!]
	\centering
		\caption{Hyperparameters for the Effect of Ensemble Size on Performance and Computation Time experiment.}\label{tab:hyperparam2}
		\begin{tabular}{l l r}
		\hline
		& Hyperparameter & Value\\
		\hline
		\emph{Problem parameters} & &\\
		& Environment & \emph{CartPole-v0}\\
		& State dimensions $|\mathcal{S}|$ & $4$\\
		& Action dimensions $|\mathcal{A}|$ & $2$\\
		& Maximum episode length & $200$\\
		\hline
		\emph{Algorithm parameters} & &\\
		& Discount $\gamma$ & $0.99$\\
		& Number of atoms & $51$\\
		& Maximum steps in env & $1e5$\\
		& Initial $\epsilon$ & $1.0$\\
		& Final $\epsilon$ & $0.05$\\
		& Samples from replay buffer & $100$\\
		& Replay buffer size & $1e5$\\
		& Minibatch size & $32$\\
		& Regularising parameter $\lambda$ & $1.0$\\
		& Return distribution range $[V_{min}, V_{max}]$ & $[0, \frac{1-\gamma^{200}}{1-\gamma}]$\\
		& Update ensembles at steps & $[100, 200, \cdots]$\\
		& Update target ensembles at steps & $[1000, 2000, \cdots]$\\
		& Ensemble size $K$ & $[1, 2, 4, 8]$\\
		& Learnable parameters & $(32 \, |\mathcal{S}| + 11891)K |\mathcal{A}|$\\
		& Optimiser & \emph{Adam}\\
		& Learning rate & $0.00025$\\
		& Network structure & $\underset{\text{input}}{4} \rightarrow \underset{\text{dense}}{32} \rightarrow \underset{\text{dense}}{32} \rightarrow \underset{\text{dense}}{128} \rightarrow \underset{\text{output}}{(51, 2)}$\\
		\hline
	\end{tabular}
\end{table*}

\clearpage
\subsection{Experiments With Heterogenous Risk Measures}
In Figure~\ref{fig:cartpole} the results are shown for experiment with different coherent risk measures. In this section the hyperparameters used for that experiment is shown.

\begin{table*}[h!]
	\centering
		\caption{Hyperparameters for the Experiments With Different Risk Measures.}\label{tab:hyperparam3}
		\begin{tabular}{l l r}
		\hline
		& Hyperparameter & Value\\
		\hline
		\emph{Problem parameters} & &\\
		& Environment & \emph{CartPole-v0}\\
		& State dimensions $|\mathcal{S}|$ & $4$\\
		& Action dimensions $|\mathcal{A}|$ & $2$\\
		& Maximum episode length & $200$\\
		\hline
		\emph{Algorithm parameters} & &\\
		& Discount $\gamma$ & $0.99$\\
		& Number of atoms & $51$\\
		& Maximum steps in env & $1e5$\\
		& Initial $\epsilon$ & $1.0$\\
		& Final $\epsilon$ & $0.05$\\
		& Samples from replay buffer & $100$\\
		& Replay buffer size & $1e5$\\
		& Minibatch size & $32$\\
		& Regularising parameter $\lambda$ & $1.0$\\
		& Return distribution range $[V_{min}, V_{max}]$ & $[0, \frac{1-\gamma^{200}}{1-\gamma}]$\\
		& Update ensembles at steps & $[100, 200, \cdots]$\\
		& Update target ensembles at steps & $[1000, 2000, \cdots]$\\
		& Ensemble size $K$ & $4$\\
        & $CVaR \circ CVaR \,(\alpha_E, \alpha_A)$ & $(\text{Epistemic }0.25, \text{Aleatory }0.25)$\\
        & $Wang \circ Wang \,(\alpha_E, \alpha_A)$ & $(\text{Epistemic }0.10, \text{Aleatory }0.10)$\\
        & $CVaR \circ sd \,(\alpha_E, \alpha_A)$ & $(\text{Epistemic }0.25, \text{Aleatory }1.0)$\\
        & $sd \circ CVaR \,(\alpha_E, \alpha_A)$ & $(\text{Epistemic }1.0, \text{Aleatory }0.25)$\\
		& Learnable parameters & $(32 \, |\mathcal{S}| + 11891)K |\mathcal{A}|$\\
		& Optimiser & \emph{Adam}\\
		& Learning rate & $0.00025$\\
		& Network structure & $\underset{\text{input}}{4} \rightarrow \underset{\text{dense}}{32} \rightarrow \underset{\text{dense}}{32} \rightarrow \underset{\text{dense}}{128} \rightarrow \underset{\text{output}}{(51, 2)}$\\
		\hline
	\end{tabular}
\end{table*}

\clearpage
\subsection{Hyperparameters for the Highway Experiment}

In Figure~\ref{fig:highway} and Table~\ref{tab:highway10} the results are shown for the highway environment. In this section the hyperparameters are shown used for that experiment.

\begin{table*}[h!]
	\centering
		\caption{Hyperparameters for the Large State Space Risk-Sensitive Experiments.}\label{tab:hyperparam4}
		\begin{tabular}{l l r}
		\hline
		& Hyperparameter & Value\\
		\hline
		\emph{Problem parameters} & &\\
		& Environment & \emph{Highway-v1}\\
		& State dimensions $|\mathcal{S}|$ & $25$\\
		& Action dimensions $|\mathcal{A}|$ & $5$\\
		& Maximum episode length & $40$\\
		& Vehicles count & $10$\\
		\hline
		\emph{Algorithm parameters} & &\\
		--\emph{Shared}& Discount $\gamma$ & $0.99$\\

		& Maximum steps in env & $1e5$\\
		& Initial $\epsilon$ & $1.0$\\
		& Final $\epsilon$ & $0.01$\\
		& Samples from replay buffer & $1000$\\
		& Replay buffer size & $1e5$\\
		& Update ensembles at steps & $[1, 3, 6, 10, \cdots]$\\
		& Update target ensembles at steps & $[15, 55, 120, \cdots]$\\
		& Ensemble size $K$ & $4$\\
		& Optimiser & \emph{Adam}\\
		& Learning rate & $0.00025$ \\
		& Minibatch size & $32$\\
		& &\\
		--\emph{SENTINEL-K} & Number of atoms & $51$\\
		& Regularising parameter $\lambda$ & $1.0$\\
		& Return distribution range $[V_{min}, V_{max}]$ & $[0, 40]$\\
		& Risk-sensitive parameters ($\alpha_E, \alpha_A)$ & $(0.25, 0.25)$\\
		& Learnable parameters & $(32 \, |\mathcal{S}| + 11891)K |\mathcal{A}|$\\
		& Network structure & $\underset{\text{input}}{25} \rightarrow \underset{\text{dense}}{32} \rightarrow \underset{\text{dense}}{32} \rightarrow \underset{\text{dense}}{128} \rightarrow \underset{\text{output}}{(51, 5)}$\\
		& &\\
		--\emph{UA-DQN} & Risk-sensitive parameters $(\beta, \lambda)$ & $(0.2, 0.1)$\\
		& Learnable parameters & $(256 \, |\mathcal{S}| + 1541)K |\mathcal{A}|$\\
		& Network structure & $\underset{\text{input}}{25} \rightarrow \underset{\text{dense}}{256} \rightarrow \underset{\text{output}}{5}$\\
		& &\\
		--\emph{VDQN-CVaR} & Risk-sensitive parameter $\alpha$ & $0.25$\\
		& Learnable parameters & $(400|\mathcal{S}|+400)|\mathcal{A}|$\\
        & Network structure & $\underset{\text{input}}{25} \rightarrow \underset{\text{dense}}{100} \rightarrow \underset{\text{output}}{5}$\\
		& &\\
		\hline
	\end{tabular}
\end{table*}

\end{document}